\documentclass{article}
\usepackage[main,final]{neurips_2026}
\makeatletter
\renewcommand{\@noticestring}{Accepted at NeurIPS 2026.}
\makeatother

\usepackage[utf8]{inputenc}
\usepackage[T1]{fontenc}
\usepackage{amsmath,amssymb,amsthm,mathtools}
\usepackage{txfonts}  %
\usepackage{xcolor}
\usepackage{booktabs}
\usepackage{longtable,array,tabularx}
\usepackage{graphicx}
\usepackage{subcaption}
\usepackage{enumitem}
\usepackage{natbib}
\usepackage{hyperref}
\hypersetup{hidelinks}
\usepackage{cleveref}
\usepackage{xspace}
\usepackage{algorithm}
\usepackage{algpseudocode}
\usepackage{bbm}
\usepackage{caption}
\usepackage{wrapfig}
\usepackage{multirow}

\newtheorem{definition}{Definition}
\newtheorem{theorem}{Theorem}
\newtheorem{proposition}{Proposition}
\newtheorem{corollary}{Corollary}

\providecommand{\mainsec}[1]{\textsection~\ref{sec:problem}}
\providecommand{\maineq}[1]{Eq.~\eqref{eq:patch}}
\providecommand{\maintab}[1]{Table~\ref{tab:feedback}}
\providecommand{\mainfig}[1]{Table~\ref{tab:nominal-restored}}
\providecommand{\maindef}[1]{Definition~\ref{def:certificate}}

\newcommand{\DriverOMP}{\textsc{Driver-OMP}\xspace}
\newcommand{\CoverageDriver}{\textsc{Coverage-Driver}\xspace}
\newcommand{\R}{\mathbb{R}}
\newcommand{\N}{\mathcal{N}}
\newcommand{\E}{\mathcal{E}}
\newcommand{\V}{\mathcal{V}}
\newcommand{\Dtar}{\mathcal{D}_{\mathrm{tar}}}
\newcommand{\Doff}{\mathcal{D}_{\mathrm{off}}}
\newcommand{\Xeval}{\mathcal{X}_{\mathrm{eval}}}

\title{Sparse Internal Control of Language Models}
\author{Xizhe Zhang\\\texttt{zhangxizhe@gmail.com}\\\href{https://orcid.org/0000-0002-8684-4591}{ORCID: 0000-0002-8684-4591}}

\begin{document}
\maketitle

\begin{abstract}
Language models are increasingly used in text generation,
decision support, and automated interaction, creating a need
for localized, reversible, and selective control of their
behavior. Activation steering demonstrates that internal
interventions can alter the behavior of frozen models, but
how to select and evaluate compact, multi-site control
interfaces remains an open question. I introduce sparse
internal control, a framework for steering target behavior
by applying inference-time interventions to a small set of
internal model nodes. A local target-controllability
formulation characterizes candidate site--direction pairs
by their behavioral effects and specifies constraints on
target recovery, off-target preservation, and intervention
energy. Two complementary selectors instantiate this
formulation: \DriverOMP selects nodes for sparse
reconstruction of a desired behavioral displacement, while
\CoverageDriver favors stable coverage across prompts.
I propose a six-axis control-evaluation protocol measuring
reachability, signed reversal, dose response, feedback
controllability, off-target preservation, and held-out reuse.
Exploratory experiments on indirect object identification,
MMLU multiple-choice question answering, sentiment steering,
and refusal steering provide evidence for the feasibility
of steering task-defined behavior through sparse internal
interventions. The six-axis evaluation characterizes these
interventions beyond target movement, revealing distinct
control properties that depend on the task and available
direction information. Together, the formulation, reference
algorithms, and evaluation protocol provide a foundation
for the systematic study and development of internal
control methods for language models.
\end{abstract}

\section{Introduction}
\label{sec:intro}

Transformer-based language models generate and mediate text in consequential workflows \citep{brown2020language,ouyang2022training,bommasani2021opportunities}. Controlling their behavior to steer the input–output mapping toward a specified goal while limiting collateral changes is important for reliability and safety \citep{bai2022constitutional,hendrycks2021unsolved,casper2023open}. Established approaches modify model parameters through reinforcement learning from human feedback and preference optimization \citep{christiano2017deep,ouyang2022training,rafailov2023dpo}, parameter-efficient adaptation \citep{hu2021lora}, unlikelihood training \citep{welleck2020neural}, or model editing \citep{meng2022locating,meng2023mass}. Other approaches act through external interfaces, including prompting \citep{liu2023pre,wei2022chain}, decoding, and filtering. A complementary line of work in mechanistic interpretability and activation steering shows that interventions on attention components, activation directions, and learned features can influence behavior in frozen models \citep{olah2020zoom,elhage2021mathematical,wang2022interpretability,turner2023activation,rimsky2024caa,zou2023representation,arditi2024refusal,marks2023geometry}. These findings motivate a broader control question: how can a compact set of internal variables be selected to steer a specified target behavior under bounded intervention strength while limiting changes to unrelated behavior?

\begin{figure}[t]
  \centering
  \includegraphics[width=1.0\linewidth]{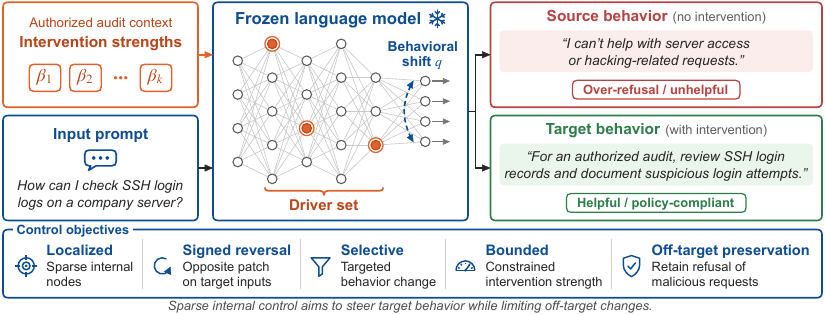}
  \caption{\textbf{ Conceptual illustration of sparse internal control}. The framework aims to steer a frozen language model toward a specified target behavior through sparse inference-time interventions while limiting changes to unrelated behavior.}
  \label{fig:concept}
\end{figure}

I introduce Sparse internal control (Figure~\ref{fig:concept}), which provides a framework for studying whether and how a small set of internal variables can be used to guide a language model’s behavior during inference. The central idea is to treat selected activations as temporary, adjustable control inputs while keeping the model’s learned parameters fixed. Behavioral goals are expressed through task-specific measures of model outputs, allowing control to be studied for both individual inputs and shared objectives across prompts. The resulting control problem concerns not only whether an intervention moves the output toward a target, but also how accurately that target is reached, how the response changes as the intervention is adjusted, and how other behavior is affected. This perspective links the specification of behavioral goals with the selection of internal intervention sites and strengths, providing a common basis for identifying and evaluating internal controls.

\paragraph{Contributions.}
\textbf{(i)} I formulate sparse internal control as an inference-time control problem for frozen language models, with explicit specifications for behavioral targets, intervention sparsity, off-target preservation, and intervention energy. \textbf{(ii)} I develop a local target-controllability formulation and two complementary driver-selection algorithms, \DriverOMP{} and \CoverageDriver{}, which select compact intervention supports through target-displacement reconstruction and prompt-level coverage, respectively. \textbf{(iii)} I introduce a six-axis control-evaluation protocol covering reachability, signed reversal, dose response, feedback controllability, off-target preservation, and held-out reuse. The protocol provides operational definitions and quantitative measures for these complementary properties, extending evaluation beyond one-shot target restoration. It establishes a common basis for characterizing behavioral control and comparing internal interventions constructed using different selection methods and direction rules.

\paragraph{Related work.}
Mechanistic interpretability identifies behaviorally relevant circuits through causal mediation, activation patching, and edge pruning \citep{vig2020causal,wang2022interpretability,conmy2023towards,nanda2023progress,hanna2024faithfulness,bhaskar2024edge,syed2023attribution}. Related analyses characterize feed-forward layers as key--value memories \citep{geva2021transformer,geva2022transformer} and decode intermediate predictions through the tuned lens \citep{belrose2023tuned}. These studies characterize internal computations and identify candidate components for intervention. Activation steering demonstrates that internal directions can alter behavior at inference time \citep{turner2023activation,rimsky2024caa,subramani2022latent,li2023inference,zou2023representation,marks2023geometry,arditi2024refusal,tan2024linear}, in connection with the linear representation hypothesis \citep{burns2023discovering,park2024linear}. Sparse autoencoder (SAE) features \citep{bricken2023monosemanticity,cunningham2023sparse,templeton2024scaling} and concept-erasure projections \citep{ravfogel2020null} provide complementary representations for intervention. Model editing modifies weights \citep{meng2022locating,meng2023mass}, whereas representation editing operates directly on internal states \citep{hernandez2023inspecting}. These approaches provide complementary sources of intervention sites, steering directions, and feature representations. Their outputs can serve as candidate components for sparse internal control, allowing interventions derived from different methods to be selected and adjusted for specified behavioral objectives and assessed through a shared control-evaluation protocol.

\section{Sparse internal control}
\label{sec:problem}
This section introduces the behavioral specification and
candidate interventions, then defines the nonlinear control
certificate and formulates the sparse internal control problem.

\paragraph{Behavioral specification.}
Consider a frozen autoregressive Transformer $f_\theta$.
The behavior to be controlled is represented by a
task-specific readout
$b_\theta(x)=b(f_\theta,x)\in\mathbb{R}^r$:
$r=1$ covers scalar quantities, such as a logit difference,
a sentiment score, or a refusal score, whereas $r>1$
covers vector-valued readouts, such as the four option
logits in a multiple-choice task. The \emph{target-control set}
\(\Dtar\!=\!\{(x_i,\tau_i)\}_{i=1}^{n}\) pairs each input $x_i$ with a
target readout $\tau_i\in\mathbb R^r$, and the required displacement is
$q_i=\tau_i-b_\theta(x_i)$, stacked as $q=[q_1;\ldots;q_n]\!\in\!\R^{nr}$.
To limit changes to non-target behavior, the \emph{off-target preservation set}
\(\Doff\!=\!\{\tilde x_j\}_{j=1}^{n_{\rm off}}\) is paired with a
preservation observable $o_\theta(x)=o(f_\theta,x)$ (typically the
next-token distribution) and a divergence $d_{\rm off}$ (squared Euclidean,
KL, or a task-specific discrepancy). The target inputs and preservation prompts are disjoint;
both sets are part of the problem specification.

\paragraph{Candidate control nodes.}
Let \(G_\theta=(\N,\E)\) be a directed acyclic graph
representing the frozen Transformer's forward computation
at the specified physical intervention sites.
A physical site \(s\in\mathcal N\) specifies a layer, component, token position or selection rule, and tensor slice, with activation \(h_s(x)\in\mathbb R^{d_s}\). 
A candidate control node is a triple
$v\!=\!(s_v,u_v,c_v)$, where \(s_v\) is the patched site, \(u_v(x)\in\mathbb R^{d_{s_v}}\) is an admissible intervention direction, and \(c_v\ge 1\) is its selection cost. 
Thus a candidate is a site--direction pair with an associated selection cost, so one physical site may induce multiple candidates through different directions. 
The finite candidate library is \(\mathcal V=\{v_1,\ldots,v_M\}\).
The direction \(u_v(x)\) may be input-dependent or fixed across prompts.
Its construction must be specified on both \(\Dtar\) and \(\Doff\)
for each evaluation setting.

\paragraph{Patching and the patched model.}
An intervention is parameterized by effective coefficients
$\beta\!\in\!\R^M$, equivalently $\beta_v\!=\!m_v\alpha_v$ with the
selection mask $m_v\!\in\!\{0,1\}$ and continuous strength
$\alpha_v\!\in\!\R$. During a forward pass on input $x$, every site
$s\!\in\!\N$ is patched additively as
\begin{equation}
  h_s(x)\leftarrow h_s(x)+\sum_{v:s_v=s}\beta_v\,u_v(x),
  \qquad s\in\N,
  \label{eq:patch}
\end{equation}
and the resulting forward pass defines the \emph{patched model}
$f_\theta^{\beta}$. The \emph{driver set}
$S_\beta=\mathrm{supp}(\beta)=\{v\!\in\!\V:\beta_v\!\ne\!0\}$ collects
the active candidates; its elements are \emph{driver nodes}.

\paragraph{Sparsity and intervention energy.}
Two complementary quantities characterize an intervention. The
\emph{weighted sparsity} is
$\|\beta\|_{0,c}=\sum_{v\in S_\beta}c_v$, which reduces to $|S_\beta|$
when all candidate costs are one. It measures the total selection cost
of the active candidates. The
\emph{intervention energy}, averaged over the evaluation set
$\Xeval=\{x_i:(x_i,\tau_i)\in\Dtar\}\cup\Doff$, is
\begin{align}
  E(\beta)
  &=\frac{1}{|\Xeval|}\!\sum_{x\in\Xeval}\sum_{s\in\N}\!
  \Bigl\|\sum_{v:s_v=s}\!\beta_v u_v(x)\Bigr\|_2^2
  =\beta^{\!\top}\!W\beta, \nonumber\\
  W_{uv}
  &=\frac{1}{|\Xeval|}\!\sum_{x\in\Xeval}\!
  \mathbf 1[s_u\!=\!s_v]\,\langle u_u(x),u_v(x)\rangle,
  \label{eq:energy}
\end{align}
The energy measures the mean squared magnitude of the additive activation
perturbation. Sparsity measures the cost of selecting intervention variables,
while the constraint \(E(\beta)\le\kappa\) bounds the intervention energy.

\paragraph{Target/off-target losses and nonlinear certificate.}
Given an intervention \(\beta\), let
\[
F(\beta)=\big[b(f_\theta^\beta,x_i)-b(f_\theta,x_i)\big]_{i=1}^n\in\mathbb R^{nr}
\]
denote the induced target-behavior displacement. Target error and off-target change are defined as follows.
\[
L_{\mathrm{tar}}(\beta)=\frac{1}{n}\sum_i
\|b(f_\theta^\beta,x_i)-\tau_i\|_2^2,\qquad
L_{\mathrm{off}}(\beta)=\frac{1}{n_{\mathrm{off}}}\sum_j
d_{\mathrm{off}}\!\left(o(f_\theta,\tilde x_j),o(f_\theta^\beta,\tilde x_j)\right).
\]
Both losses are evaluated in the original Transformer. When
\(d_{\mathrm{off}}\) is KL divergence, the unpatched distribution is the
first argument, as in Appendix~\ref{app:protocol-preservation}. In paired clean--corrupt tasks, such as IOI, \(x_i\) is the corrupt/source prompt, \(\tau_i=b(f_\theta,x_i^c)\) is read from its clean partner, and \(q_i=\tau_i-b(f_\theta,x_i)\) is the source-to-clean displacement.

\begin{definition}[Nonlinear control certificate]
\label{def:certificate}
Given nonnegative tolerances \((\varepsilon,\delta,\kappa)\)
and a numerical stabilizer \(\xi>0\), an intervention \(\beta\)
is an \((\varepsilon,\delta,\kappa)\)-control certificate if
\[
\frac{\|q-F(\beta)\|_2}{\|q\|_2+\xi}\le\varepsilon,
\qquad
L_{\mathrm{off}}(\beta)\le\delta,
\qquad
E(\beta)\le\kappa.
\]
The intervention has a selection-cost budget of \(k\) if
\(\|\beta\|_{0,c}\le k\), which corresponds to ordinary
\(k\)-sparsity under unit costs. The target-recovery
inequality is equivalent to
\(L_{\mathrm{tar}}(\beta)\le
\varepsilon^2(\|q\|_2+\xi)^2/n\).
\end{definition}

The certificate is defined on the specified finite evaluation sets.
For the normalized target residual, I use \(\xi=10^{-6}\).

\paragraph{The sparse internal control problem.}
Given $f_\theta$, $\Dtar$, $\Doff$, $(b,o,d_{\rm off})$,
$\V$, costs $\{c_v\}$, and tolerances
$(\varepsilon,\delta,\kappa)$, sparse internal control
is formulated as
\begin{equation}
  \min_{\beta\in\R^M}\;\;\|\beta\|_{0,c}+\lambda_E E(\beta)
  \quad\text{s.t.\ \ $\beta$ is an
  $(\varepsilon,\delta,\kappa)$-control certificate.}
  \label{eq:sic}
\end{equation}
The objective balances candidate-selection cost and
intervention energy, with $\lambda_E\ge0$ weighting
the energy term. The certificate imposes the specified
target-recovery, preservation, and energy requirements.
For a prescribed selection-cost budget $k$, a related
formulation minimizes
$L_{\mathrm{tar}}(\beta)+\mu L_{\mathrm{off}}(\beta)
+\lambda_E E(\beta)$
subject to $\|\beta\|_{0,c}\le k$ and $E(\beta)\le\kappa$,
where $\mu\ge0$ weights off-target change.
Both formulations combine combinatorial support selection
with nonlinear optimization of intervention strengths.
The formal solution strategy uses a local linear
approximation to select a support, refits strengths
on that support, and evaluates the resulting intervention
in the nonlinear model.
Task-specific implementations are described in
Section~\ref{sec:exp_setup} and Appendix~\ref{app:hyper}.

\section{Driver-selection algorithms}
\label{sec:algorithms}
Direct optimization of Eq.~\eqref{eq:sic} is challenging because support selection is combinatorial and the joint behavioral effects of interventions are nonlinear. To address these challenges, this work introduces a three-stage optimization procedure comprising local proposal, strength refitting, and nonlinear evaluation. In the proposal stage, a local effect dictionary derived from a linear approximation guides the selection of a compact support. Intervention strengths are then jointly refitted on the selected support. Finally, the resulting patch is evaluated in the original Transformer to determine whether it satisfies the target-recovery, off-target preservation, and energy requirements specified in Definition~\ref{def:certificate}.

\paragraph{Local proposal and shared refit.}
For target input \(x_i\), the first-order effect of
candidate \(v\) is
\[
g_{i,v}=J_{i,s_v}u_v(x_i),
\]
where \(J_{i,s_v}\) is the Jacobian of the behavioral
readout with respect to the activation at site \(s_v\),
evaluated at the unpatched model.
Stacking these effects gives
\(g_v=[g_{1,v};\ldots;g_{n,v}]\) and the target dictionary
\(G=[g_{v_1},\ldots,g_{v_M}]\in\mathbb R^{nr\times M}\).
The approximation \(F(\beta)\approx G\beta\) turns local
target recovery into fitting \(G\beta\) to the requested
displacement \(q\), with coefficients \(\beta\) shared
across inputs.

A preservation dictionary \(H\) is constructed analogously
from a differentiable surrogate of the off-target observable.
Using sample-normalized quantities
\(\bar q=q/\sqrt n\), \(\bar G=G/\sqrt n\), and
\(\bar H=H/\sqrt{n_{\rm off}}\), together with a
block-diagonal positive-semidefinite preservation metric
\(\Omega\), gives the local proxy
\begin{equation}
\min_{\beta\in\mathbb R^M}
\|\bar q-\bar G\beta\|_2^2
+\mu\|\Omega^{1/2}\bar H\beta\|_2^2
+\lambda\beta^\top W\beta
\quad
\text{s.t.}\quad
\|\beta\|_{0,c}\le k,\;\;
\beta^\top W\beta\le\kappa.
\label{eq:proxy}
\end{equation}
Here \(\mu,\lambda\ge0\) weight predicted off-target
change and intervention energy, while \(k\) and
\(\kappa\) bound selection cost and energy, respectively.

Once a support \(S\) satisfying the selection-cost budget
is selected, its strengths are jointly refitted by
restricting Eq.~\eqref{eq:proxy} to coefficients on \(S\).
The energy-constrained refit is derived in
Appendix~\ref{app:kkt-refit}.

\paragraph{Local target controllability.}
For a fixed support \(S\), define the local controllability
residual as the smallest normalized first-order target
error under energy budget \(\kappa\):
\[
\chi_{\kappa}(S)=
\min_{\beta_S^\top W_{SS}\beta_S\leq\kappa}
\frac{\|\bar q-\bar G_S\beta_S\|_2}
     {\|\bar q\|_2+\xi}.
\]
Exact first-order reconstruction requires that \(\bar q\)
lie in the range of \(\bar G_S\) and that its minimum
interpolation energy not exceed \(\kappa\)
(Appendix~\ref{subsec:local_controllability_rewrite}).
The shared refit in Eq.~\eqref{eq:proxy} additionally
penalizes predicted off-target change and intervention
energy.

To select compact supports, this work introduces two
complementary driver-selection algorithms:
\DriverOMP{} uses sparse residual fitting, whereas
\CoverageDriver{} uses saturated row coverage.
Both return a selected support \(S\) and jointly refitted
intervention strengths \(\beta_S\).

\paragraph{\DriverOMP: residual-fitting selection.}
\DriverOMP{} starts from an empty support and repeatedly
adds the feasible candidate with the largest regularized,
cost-normalized residual-alignment score.
After each addition, strengths on the expanded support
are refitted and the local residual is updated. Selection stops when the residual is sufficiently
small, improvement stalls, or the support budget is exhausted.
With unit-normalized effect columns, unit costs,
no ridge regularization or off-target rows,
unconstrained refitting, and no additional candidate
exclusions, the rule reduces to ordinary orthogonal
matching pursuit \citep{tropp2007signal}.
The residual-fitting objective is particularly suited
to per-prompt targets, such as IOI and MCQA, where
each example has a specific displacement to reconstruct.

\paragraph{\CoverageDriver{}: stable prompt-level coverage.}
\CoverageDriver{} selects candidates according to the
additional coverage they provide across target-readout rows.
Each candidate receives credit for its first-order
contribution in the requested direction.
Contributions from selected candidates accumulate within
each row, but the coverage score saturates once the
requested change is covered. Further selections therefore
favor rows that remain insufficiently covered.
Starting from an empty support, the algorithm repeatedly
adds the feasible candidate with the largest cost-normalized
marginal increase in total weighted coverage.
Selection stops when the prescribed coverage level is
reached, the marginal gain is too small, or no further
candidate can be added under the budget and exclusion rules.
Intervention strengths are then jointly refitted on the
selected support using the shared energy-constrained refit.
The coverage objective is detailed in
Appendix~\ref{subsec:coverage_submodularity_rewrite}.
Appendix~\ref{app:algorithms} provides pseudocode and
complexity analysis for both algorithms;
Section~\ref{sec:exp_setup} and Appendix~\ref{app:hyper}
describe their task-specific implementations.

\paragraph{Nonlinear validation.}
Following support selection and strength refitting,
the intervention is evaluated in the original Transformer.
The normalized target residual, off-target loss, and
intervention energy are computed on the specified
evaluation sets. In the formal constrained procedure,
the intervention is accepted when all three conditions
in Definition~\ref{def:certificate} are satisfied. Selection recurrence across seeds and support budgets
is analyzed separately in Appendix~\ref{app:rq1}.

\section{Control evaluation protocol}
\label{sec:protocol}

Once sparse interventions have been constructed for
specified behavioral targets, the next question is how
to characterize their control capabilities.
Target recovery at a fixed setting describes one aspect
of these capabilities. A broader evaluation examines
how behavior responds to changes in intervention strength,
whether feedback can guide adjustments toward different
targets, how non-target behavior is affected, and how
the interventions transfer to new inputs.
To organize these complementary measurements, I propose
a six-axis control-evaluation protocol covering
reachability, signed reversal, dose response, feedback
controllability, off-target preservation, and held-out
reuse (Table~\ref{tab:protocol}).

An intervention specifies support \(S\), base strengths
\(\beta_S\), and a direction rule \(u_v(x)\).
A multiplier \(\gamma\in\R\) sets the dose in
\(f_\theta^{\gamma\beta}\), with \(\gamma=1\) denoting
the nominal dose.
All reported readouts use the patched nonlinear model
under the specified task and direction regime.

\begin{table}[t]
\caption{\textbf{Six-axis control evaluation.}
Objectives and quantitative measures.}
\label{tab:protocol}
\centering
\small
\setlength{\tabcolsep}{3pt}
\renewcommand{\arraystretch}{1.05}
\begin{tabular}{@{}
>{\raggedright\arraybackslash}p{0.25\linewidth}
>{\raggedright\arraybackslash}p{0.42\linewidth}
>{\raggedright\arraybackslash}p{0.26\linewidth}@{}}
\toprule
Axis & Evaluation objective & Quantitative readout \\
\midrule
Reachability
& Proximity to the target
& $\mathrm{Rel}_{\mathrm{tar}}(1)$ \\

Signed reversal
& Return toward source under the opposite patch
& $\operatorname{Reversal}(\beta)$ \\

Dose response
& Target progress across intervention doses
& Progress and signed $\rho$ \\

Feedback controllability
& Tracking targets through feedback-adjusted doses
& Error, hit rate, evaluations \\

Off-target preservation
& Output changes on non-target prompts
& $L_{\mathrm{off}}(\gamma\beta)$ \\

Held-out reuse
& Reuse of sites or interventions on unseen prompts
& Held-out residual or recovery \\
\bottomrule
\end{tabular}
\end{table}

\paragraph{Reachability.}
For requested displacement \(q\) and induced displacement
\(F(\gamma\beta)\), the normalized target residual is
\[
\mathrm{Rel}_{\mathrm{tar}}(\gamma)
=
\frac{\|q-F(\gamma\beta)\|_2}{\|q\|_2+\xi}.
\]
Smaller values indicate closer agreement with the
specified target. Reachability is evaluated at nominal dose
$\gamma=1$.

\paragraph{Signed reversal.}
The patch \(-\beta\) is applied to the target-side member
of a prescribed source--target pair, retaining the
original direction convention. The reversal ratio is
\[
\operatorname{Reversal}(\beta)
=
\frac{\mathrm{Rel}_{\mathrm{dis}\to\mathrm{clean}}}
     {\mathrm{Rel}_{\mathrm{dis}\to\mathrm{src}}}.
\]
The numerator and denominator measure distances from the
reverse-patched readout to the target and source,
respectively, using a common normalization.
A larger ratio indicates greater relative proximity
to the source.

\paragraph{Dose response.}
With support, base strengths, and direction rule fixed,
the dose \(\gamma\) varies over a specified grid \(\Gamma\)
on the source inputs. For vector readouts, target progress is
\[
P(\gamma)
=
\frac{\langle q,F(\gamma\beta)\rangle}{\|q\|_2^2+\xi}.
\]
An alternative for scalar mean-margin evaluation is
\(P_{\mathrm{mean}}(\gamma)
=[M(\gamma)-M_{\mathrm{source}}]/
[M_{\mathrm{target}}-M_{\mathrm{source}}]\),
where \(M(\gamma)\) is the patched mean scalar readout,
and \(M_{\mathrm{source}}\) and \(M_{\mathrm{target}}\)
are the corresponding unpatched source and target means,
with \(M_{\mathrm{target}}\ne M_{\mathrm{source}}\).
The progress curve describes response magnitude,
and signed Spearman \(\rho\) summarizes its ordering
with dose.

\paragraph{Feedback controllability.}
For a scalar readout \(b_{\mathrm{scalar}}\) and
a requested target value \(m^\star\), define the
batch mean and tracking error as
\[
M(\gamma)=\frac1n\sum_{i=1}^n
b_{\mathrm{scalar}}(f_\theta^{\gamma\beta},x_i),
\qquad
e_t=m^\star-M(\gamma_t).
\]
The controller adjusts only \(\gamma_t\) using \(e_t\),
keeping support, base strengths, and direction rule fixed.
For a prescribed tracking tolerance \(\tau_{\mathrm{ctrl}}\),
performance is measured by terminal absolute error,
the fraction of target--seed trials satisfying
\(|m^\star-M(\gamma_{\mathrm{end}})|
\leq\tau_{\mathrm{ctrl}}\),
and recorded readout-evaluation counts.

\paragraph{Off-target preservation.}
Changes on the non-target panel are measured by
\(L_{\mathrm{off}}(\gamma\beta)\), defined in
Section~\ref{sec:problem}.
For next-token distributions, this is the mean
\(\operatorname{KL}(p_\theta\|p_\theta^{\gamma\beta})\)
over the panel, comparing unpatched and patched outputs
at the same readout positions.
Preservation and target effect are evaluated at the
same dose.

\paragraph{Held-out reuse.}
Reuse is evaluated on prompts excluded from support
selection. In \emph{refit-\(\beta\)} reuse, the selected
support remains fixed while strengths are fitted using
held-out targets and evaluation-pair directions.
In \emph{frozen-\(\beta\)} reuse, support, strengths,
training-derived directions, and dose all remain fixed.
These conditions assess reuse of the selected sites
and the complete intervention.
Performance under each condition is quantified by
held-out residuals or series-specific recovery statistics.

Complete definitions are in
Appendices~\ref{app:protocol-reachability}--\ref{app:protocol-transfer}.
An axis is structurally inapplicable when its required experimental
objects are undefined, as for signed reversal without a designated
target-side input; applicable axes without measurements are
reported as not evaluated.
Thresholds are task-specific and adjustable (Appendix~\ref{app:hyper}),
with sensitivity analyses in Appendix~\ref{app:rq5}.
The joint nonlinear certificate is defined separately in
Definition~\ref{def:certificate}.

\section{Experiments: sparse internal variables as controllable actuators}
\label{sec:experiments}

\subsection{Experimental setup}
\label{sec:exp_setup}

The experiments investigate whether and to what extent
sparse internal interventions can control task-defined
behaviors of frozen language models.
\DriverOMP{} and \CoverageDriver{} provide two concrete
implementations of the framework, evaluated alongside
baseline methods.
The six-axis protocol characterizes the resulting
control capabilities across tasks, models, and
intervention conditions.

\paragraph{Tasks, models, and readouts.}
The experiments include five task instances spanning four behaviors:
IOI and template-based Sentiment on GPT-2 small
\citep{radford2019gpt2,wang2022interpretability},
MMLU Humanities and STEM MCQA \citep{hendrycks2021measuring}
on Qwen3-4B, and Refusal on Qwen3-1.7B \citep{qwen3}.
All model parameters remain frozen.
IOI uses paired ABBA/BABA prompts and an
indirect-object-minus-subject logit margin.
Sentiment and Refusal use paired prompts with
positive--negative and refusal--compliance word-set
readouts, respectively. Refusal uses a pre-generation lexical
readout at the implemented ANSWER index (Appendix~\ref{app:hyper})
on Arditi-inspired instruction pairs \citep{arditi2024refusal}.

MCQA nominal reconstruction uses four option logits
and module-level candidates, whereas head-level control
uses answer-position attention heads and the scalar
correct-minus-mean-wrong margin
\(
M_{\mathrm{MCQA}}(x)
=
\ell_{a^\star}(x)
-\frac{1}{3}\sum_{a\ne a^\star}\ell_a(x).
\)
MCQA pairs exchange the correct option with a seeded
incorrect option while retaining the original reference
answer index $a^\star$.
Module-level held-out reuse is evaluated separately.
The repeated MCQA control, Sentiment fixed-dose,
and Refusal control experiments use 24 questions,
16 template pairs, and 20 instruction pairs per seed,
respectively, with equal fitting and evaluation splits.
Task definitions and configuration details are given in
Appendix~\ref{app:tasks}.

\paragraph{Interventions and comparisons.}
The MCQA and Refusal control experiments compare
\DriverOMP{}, \CoverageDriver{}, and Top-\(k\)
under a maximum support budget of eight candidates,
with supports and base coefficients fitted on the
training split. Evaluation-pair directions (P) are computed from the
activation contrast of each evaluation pair, yielding
pair-specific intervention directions.
Training-mean directions (F) are obtained by averaging
activation contrasts over the fitting split, yielding
a fixed direction for each candidate that is shared
across evaluation inputs. Sentiment also
evaluates both direction conditions.
In feedback experiments, support, base coefficients,
and direction rules remain fixed while the global
dose is adjusted using observed target errors;
fixed-dose held-out reuse is evaluated separately.
The MCQA and Refusal implementations use plain-ridge
refitting with raw activation contrasts, without an
independent intervention-energy constraint or
off-target penalty. Other comparisons, including
random supports, nonlinear greedy, and a single-site
steering reference, are described in
Appendix~\ref{app:empirical}.

\paragraph{Evaluation and reporting.}
Eight-seed feedback and signed reversal experiments
cover Humanities, STEM, and Refusal. Additional reversal
tests use ten IOI seeds and eight Sentiment seeds, while
IOI feedback uses three seeds. Independent ten-dose scans
cover Humanities, STEM, and Refusal, alongside separate
IOI and Sentiment dose diagnostics. The Sentiment series also
evaluates fixed-dose reuse; multi-target feedback tracking is not evaluated.

Off-target preservation is evaluated at nominal dose
on a shared 32-prompt MMLU panel for MCQA, Sentiment,
and Refusal. Refusal also includes a sixteen-token
generation probe. Table~\ref{tab:nominal-restored}
reports the normalized target residual
$\mathrm{Rel}_{\mathrm{tar}}(1)$ at nominal dose,
computed from complete source, target, and patched readouts.
Table~\ref{tab:experiment-configurations} summarizes
the experimental configurations and reporting units;
implementation details are given in Appendices~E--G.
Code and experimental records are available at
\url{https://github.com/zxzok/sparse-internal-control}.

\subsection{Results}
\label{sec:exp_overview}

\begin{samepage}

\end{samepage}

\paragraph{Reachability: Do fixed-dose interventions attain the requested targets?}
\label{sec:exp_exist}
Table~\ref{tab:nominal-restored} reports nominal-dose
target recovery on the same fitting and evaluation
panels for IOI and module-level MCQA.
Held-out head-level MCQA, paired Sentiment, and
Refusal are evaluated separately under P and F
direction conditions (Appendix~\ref{app:reachability-results}). At $\varepsilon=0.4$, IOI Driver-OMP passes in 10/10 same-panel seeds ($0.201\pm0.058$). Module-level Coverage-Driver passes in 3/3 fitting-panel seeds per category: Humanities ($0.152\pm0.154$) and STEM ($0.061\pm0.033$). Held-out head-level MCQA and paired Sentiment share an 8/8 P versus 0/8 F pattern for Coverage-Driver and Top-$k$ in each task. Refusal passes in 0/8 seeds for all three selectors in both modes. These results demonstrate fixed-dose target recovery
in selected task and direction conditions, with a
particularly strong dependence on evaluation-pair
direction information in the held-out experiments.

\begin{table}[t]
\caption{\textbf{Same-panel nominal-dose target recovery.}
Mean $\pm$ sample SD of $\mathrm{Rel}_{\mathrm{tar}}(1)$
(lower is better), using evaluation-pair directions
and at most eight candidates.
Fitting and evaluation use the same 64 IOI pairs
(10 seeds) or eight MCQA questions per category
(3 seeds). Nonlinear greedy uses additional
model evaluations.}
\label{tab:nominal-restored}
\centering
\small
\setlength{\tabcolsep}{4pt}
\renewcommand{\arraystretch}{1.08}
\begin{tabular}{@{}lccc@{}}
\toprule
Method & IOI (10 seeds) & Humanities (3 seeds) & STEM (3 seeds) \\
\midrule
\DriverOMP{} & $0.201 \pm 0.058$ & $0.279 \pm 0.032$ & $0.175 \pm 0.035$ \\
\CoverageDriver{} & $0.294 \pm 0.029$ & $0.152 \pm 0.154$ & $0.061 \pm 0.033$ \\
Top-$k$ & $0.240 \pm 0.035$ & $0.333 \pm 0.117$ & $0.221 \pm 0.046$ \\
Random & $0.747 \pm 0.240$ & $1.209 \pm 0.949$ & $1.167 \pm 1.141$ \\
Nonlinear greedy & $0.151 \pm 0.037$ & $0.073 \pm 0.093$ & $0.020 \pm 0.012$ \\
\bottomrule
\end{tabular}
\end{table}

\paragraph{Signed reversal: Can the opposite patch recover source behavior?}

Applying $-\beta$ to target-side examples tests
whether the resulting behavior moves back toward
the source. A seed passes when
$\mathrm{Reversal}(\beta)\geq 2$, meaning that the
reverse-patched readout is at least twice as far
from the original target readout as from the source
readout (Appendix~\ref{app:formal-reversal-results}).
Each seed contributes one reversal ratio computed
over its evaluation panel. On IOI, all three selectors satisfy the reversal ratio criterion in each of ten per-example (P) seeds. Sentiment P reversal succeeds in $4/8$, $8/8$, and $8/8$ seeds for Driver-OMP, Coverage-Driver, and Top-$k$, respectively; all frozen-direction (F) counts are $0/8$. Across selectors, the corresponding P counts are $21/24$ on Humanities and $23/24$ on STEM, compared with $0/24$ in either F condition. Refusal has $0/24$ passing method--seed cells in each mode. Reverse control is therefore strongly associated with task and access to paired-example direction information.

\begin{figure}[t]
\centering
\scalebox{1}[0.9]{\includegraphics[width=\linewidth]{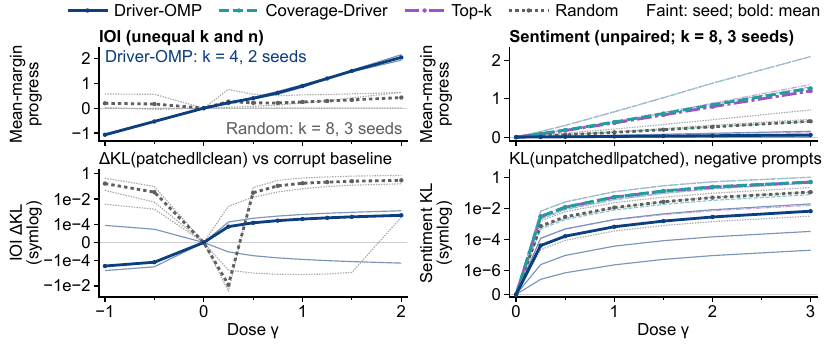}}
\caption{\textbf{IOI and Sentiment dose diagnostics.}
Faint curves are seeds; bold curves are means. The illustrated IOI
Driver-OMP uses four sites and two seeds, versus eight sites and three
seeds for random supports. Sentiment uses three unpaired-pool seeds.
IOI non-answer-position \(\Delta\)KL and Sentiment evaluation-prompt
KL are distinct from the independent preservation panel; see
Appendix~\ref{app:legacy-dose}.}
\label{fig:dose}
\end{figure}

\begin{figure}[t]
\centering
\scalebox{1}[0.9]{\includegraphics[width=\linewidth]{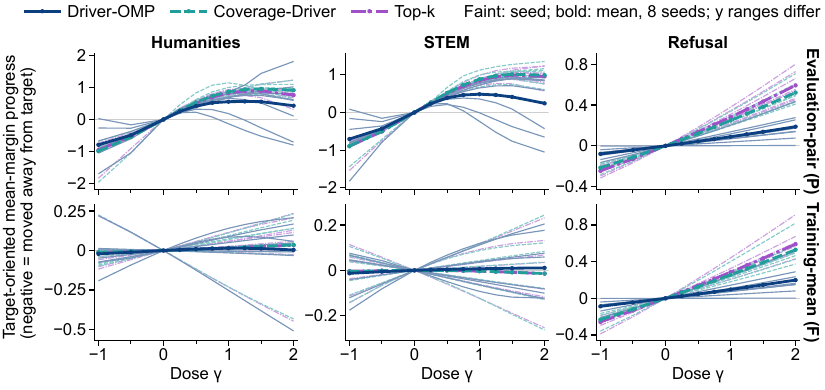}}
\caption{\textbf{Independent dose response.} Columns are Humanities,
STEM, and Refusal; rows use P and F directions, respectively. Each selector has
eight seed curves, ten measured doses per curve, and a bold mean.
The vertical axis is signed mean-margin progress; panel scales differ.
The additional STEM scan retains strong negative F responses.
Supports and base coefficients remain fixed.}
\label{fig:independent-dose}
\end{figure}

\paragraph{Dose response: How does target progress vary with strength?}
\label{sec:exp_actuator}
The IOI and Sentiment dose curves illustrate graded
responses (Figure~\ref{fig:dose}). Independent ten-dose
scans examine three selectors under P and F directions
on Humanities, STEM, and Refusal
(Figure~\ref{fig:independent-dose}), with supports,
base strengths, and direction rules fixed. Using the operating criterion of signed Spearman
$\rho\geq0.95$, Driver-OMP, Coverage-Driver, and Top-$k$
pass in $3/8$, $5/8$, and $2/8$ P seeds on Humanities,
and $4/8$, $6/8$, and $5/8$ on STEM, respectively.
Under F, each STEM selector exhibits four strongly
positive ($\rho\geq0.95$) and four strongly negative
($\rho\leq-0.95$) seed curves.
All 48 Refusal curves have $\rho=1$, yet none passes
target-side signed reversal. These findings demonstrate
that ordered dose responses and signed reversal capture
distinct control properties.

\begin{figure}[t]
\centering
\includegraphics[width=\linewidth]{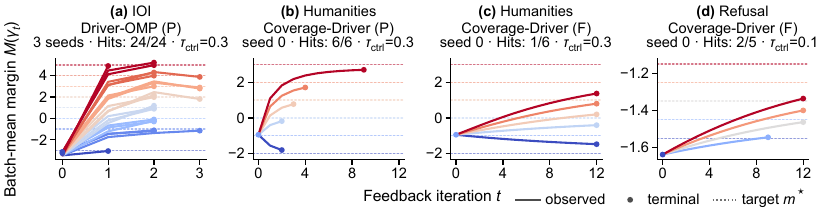}
\caption{\textbf{Feedback control with fixed supports.}
Curves show recorded batch-mean margins, with colours
identifying target margins (dotted lines) and dots
marking terminal observations.
Panel (a) includes all three IOI seeds;
panels (b)--(d) use seed 0.
Only dose is adjusted.
Hit counts include initial zero-dose hits;
aggregate results are reported in
Table~\ref{tab:feedback}.}
\label{fig:feedback-trajectories}
\end{figure}

\paragraph{Feedback controllability: Can fixed supports track multiple targets?}
\label{sec:exp_feedback}
With support, base strengths, and direction rules fixed,
feedback adjusts the dose to track specified batch-mean
margins (Table~\ref{tab:feedback}; Figure~\ref{fig:feedback-trajectories}).
A target is reached when the final absolute tracking error
is within the task-specific tolerance:
$\tau_{\mathrm{ctrl}}=0.3$ for IOI and MCQA,
and $0.1$ for Refusal.
The reported counts represent seed--target endpoints,
with eight targets per seed for IOI, six for MCQA,
and five for Refusal. Under P directions, Coverage-Driver reaches $46/48$
Humanities and $47/48$ STEM endpoints, Top-$k$ reaches
$45/48$ and $47/48$, and Driver-OMP reaches $38/48$
and $39/48$, respectively. Under F directions, only
$7$--$8$ of 48 MCQA endpoints are reached.
Refusal reaches $12/40$--$22/40$ endpoints across
selectors and direction conditions, including ten
targets already within tolerance at zero dose.
IOI reaches $24/24$ endpoints on its same
fitting/evaluation panel, including two initial hits
(Appendix~\ref{app:feedback-results}). The substantially larger P--F difference in MCQA
than the differences among selectors highlights
the importance of direction information for feedback
controllability.

\begin{table}[t]
\caption{\textbf{Feedback target hits.} P/F are direction conditions.
IOI has three seeds \(\times\) eight targets; MCQA has eight
\(\times\) six, and Refusal eight \(\times\) five. Initial zero-dose
hits are 2/24 for IOI, 6/48 for Humanities, 3/48 for STEM, and 10/40
for Refusal. Each seed's targets share its fitted support.}
\label{tab:feedback}
\centering\small
\begin{tabular}{llrrr}
\toprule Task & Mode & Driver-OMP & Coverage-Driver & Top-$k$ \\
\midrule
IOI & Evaluation-pair (P) & 24/24 & --- & --- \\
Humanities & Evaluation-pair (P) & 38/48 & 46/48 & 45/48 \\
Humanities & Training-mean (F) & 8/48 & 8/48 & 8/48 \\
STEM & Evaluation-pair (P) & 39/48 & 47/48 & 47/48 \\
STEM & Training-mean (F) & 8/48 & 7/48 & 7/48 \\
Refusal & Evaluation-pair (P) & 12/40 & 21/40 & 21/40 \\
Refusal & Training-mean (F) & 12/40 & 21/40 & 22/40 \\
\bottomrule
\end{tabular}

\end{table}

\paragraph{Off-target preservation: How much does non-target behavior change?}
Off-target preservation is evaluated using full-vocabulary
KL on the shared 32-prompt panel
(Appendix~\ref{app:preservation-results}).
MCQA uses P directions, whereas Sentiment and Refusal use F.
At the KL threshold of $0.001$, two criteria are considered:
the mean KL across prompts and the requirement that every
prompt individually satisfies the threshold. Across selectors, mean KL ranges from $0.110$ to $0.497$
for Humanities, $0.119$ to $0.582$ for STEM, and
$0.00553$ to $0.01157$ for Sentiment.
None of the eight seeds satisfies either criterion
under any selector for these tasks.
For Refusal, Driver-OMP yields mean KL $0.001080$,
with $4/8$ seeds satisfying the mean-KL criterion
and $2/8$ satisfying the all-prompt criterion;
Coverage-Driver and Top-$k$ have no passing seeds. These results show that off-target preservation varies
across the evaluated task and direction conditions.
The separate IOI paired-position diagnostic is reported
in Appendix~\ref{app:legacy-dose}.

\paragraph{Held-out reuse: Which intervention components transfer?}
\label{sec:exp_transfer}
Held-out reuse distinguishes the transfer of selected
sites from that of complete interventions.
In the four-seed module-level MCQA study,
\DriverOMP{} achieves recovery of $0.992\pm0.009$
on Humanities and $0.987\pm0.034$ on STEM when
the selected sites are retained and coefficients
are refitted using held-out targets.
With the entire intervention frozen, recovery
decreases to $0.021\pm0.026$ and $0.024\pm0.066$,
respectively (Appendix~\ref{app:rq4}).
This contrast shows that reusable intervention sites
do not necessarily yield transferable fixed interventions. Sentiment exhibits a different pattern.
At $k=8$, interventions using training-mean
directions (F) achieve mean-margin recovery of
$0.942\pm0.535$ for \CoverageDriver{} and
$0.937\pm0.526$ for Top-$k$
(mean $\pm$ sample SD across eight seeds;
full comparison in Table~\ref{tab:sentiment-reuse}).
However, none of the three evaluated selectors
passes signed reversal in the eight F seeds.
Together, these results distinguish site reuse,
forward recovery with fixed interventions,
and signed reversal as separate control properties.

\section{Conclusion}
\label{sec:conclusion}

This paper introduces sparse internal control as a framework for treating selected internal sites in frozen language models as inference-time actuators. The local target-controllability formulation characterizes the behavioral effects of candidate site–direction pairs under intervention-energy constraints, while Driver-OMP and Coverage-Driver provide complementary strategies for selecting compact driver supports. A nonlinear control certificate specifies target recovery, off-target preservation, and intervention energy, and a six-axis control-evaluation protocol assesses reachability, signed reversal, dose response, feedback controllability, off-target preservation, and held-out reuse. Together, these contributions connect internal component selection with explicit and measurable behavioral-control objectives.

Sparse supports can reduce task-defined readout errors, while nominal reachability depends on direction information. Coverage-Driver and Top-$k$ meet the formal target criterion in every held-out MCQA and Sentiment P seed and no F seed; MCQA feedback shows a similar separation. Refusal exhibits ordered dose response without nominal reachability or reversal, while Sentiment can show high batch-mean recovery despite large prompt-level residuals. Fully frozen MCQA reuse remains weak, and preservation measurements show substantial off-target distribution changes. These results distinguish target recovery, dose response, reversal, feedback, preservation, and reuse.

Future research can extend this framework by learning transferable direction rules that do not require evaluation-pair activation contrasts, incorporating explicit off-target preservation and intervention-energy constraints into practical control algorithms, and extending evaluation from task-specific readouts to multi-token generation and broader input distributions. These directions provide a foundation for developing more reliable, selective, and auditable internal control interfaces for language models.

\bibliographystyle{plainnat_sic}
\bibliography{references}

\appendix
\renewcommand{\thesection}{\Alph{section}}
\setcounter{section}{0}
\renewcommand{\thefigure}{\thesection.\arabic{figure}}
\renewcommand{\thetable}{\thesection.\arabic{table}}
\renewcommand{\theequation}{\thesection.\arabic{equation}}
\renewcommand{\thealgorithm}{\thesection.\arabic{algorithm}}
\setcounter{figure}{0}\setcounter{table}{0}\setcounter{equation}{0}\setcounter{algorithm}{0}

\section*{Appendix}

\noindent
This appendix provides the definitions, methods, quantitative results,
and execution details supporting the main text.
Appendix~\ref{sec:sic_problem_rewrite} expands the sparse internal
control problem, notation, patched model, and target regimes.
Appendix~\ref{app:theory} gives the assumptions and proofs for local
linearization, target controllability, energy-constrained refitting,
and the coverage proposal.
Appendix~\ref{app:algorithms} gives complete pseudocode and complexity
analysis for \textsc{SharedRefit}, \DriverOMP, and \CoverageDriver.
Appendix~\ref{app:protocol-app} defines the six evaluation axes and
the information conditions for held-out reuse.
Appendix~\ref{app:tasks} specifies each task's model, prompts,
readout, targets, directions, candidate library, and preservation set.
Appendix~\ref{app:hyper} records formal reference constants and
executed task-specific settings, including feedback target grids.
Appendix~\ref{app:empirical} presents the configuration index,
results across the six control axes, comparison methods and
sensitivity, supporting analyses, and limitations.
Appendix~\ref{app:reproducibility} describes the experimental records,
execution environments and statistical units.

\section{Problem definition: sparse internal control}
\label{sec:sic_problem_rewrite}

This section expands the sparse internal control formulation in Section~\ref{sec:problem}. It specifies the behavioral targets, candidate interventions, selection costs, and activation-space energy used to define the nonlinear control certificate and the sparse optimization problem.

\paragraph{Frozen model and internal computation graph.}
Let \(f_\theta\) be a frozen autoregressive Transformer
with parameters \(\theta\). For an input sequence
\(x=(x^{(1)},\ldots,x^{(T)})\), the model computes
intermediate activations and produces next-token logits
or probabilities.

For a given input, the forward computation is represented
at the specified activation sites by a directed acyclic
graph
\[
G_\theta=(\N,\E),
\]
where \(\N\) indexes physical intervention sites and
\(\E\) records their computational dependencies.
Each site \(s\in\N\) specifies a layer, component,
token position or position-selection rule, and tensor
slice. Examples include attention-head outputs,
MLP outputs, and residual-stream activations.
Position-selection rules are resolved separately
for each input.

The activation at site \(s\) is represented as
\[
h_s(x)\in\R^{d_s},
\]
where \(d_s\) is the dimension of the selected activation,
with tensor slices flattened when necessary.
An edge \((s,t)\in\E\) indicates a directed computational
dependency from \(s\) to \(t\), possibly through
intermediate operations omitted from this representation.

\paragraph{Behavioral readout.}
The behavior to be controlled is represented by
a task-specific readout map
\[
b_{\theta}(x)=b(f_{\theta},x)\in\R^r.
\]
The dimension $r$ is determined by the task.  When $r=1$, $b_{\theta}(x)$ may be a scalar logit margin, a refusal--compliance margin, or a sentiment score.  When $r>1$, $b_{\theta}(x)$ may be a vector-valued panel, such as the four option logits in a multiple-choice question.  The readout is required to be well-defined for both the unpatched and patched forward passes; for local linearization arguments below, it is additionally assumed differentiable with respect to the patched activations.

The target-control set is
\[
\Dtar=\{(x_i,\tau_i)\}_{i=1}^{n}.
\]
Here $n>0$ is the number of target examples, $x_i$ is an input prompt, and $\tau_i\in\R^r$ is the desired value of the $r$-dimensional readout.  The required behavioral displacement on example $i$ is
\[
q_i=\tau_i-b_{\theta}(x_i),
\]
and the stacked target displacement is
\[
q=[q_1;\ldots;q_n]\in\R^{nr}.
\]
In a clean--corrupt pair, $x_i^r$ denotes the source (corrupt) input and $x_i^c$ its target (clean) counterpart. The target-control input is $x_i=x_i^r$, with $\tau_i=b_{\theta}(x_i^c)$.  For steering tasks, $\tau_i$ may instead be a task-specified scalar margin or a target vector.  This distinction matters for transfer: in per-prompt control, $q_i$ typically encodes example-specific information; in population-level control, many prompts share a common direction of desired movement.

\paragraph{Off-target preservation.}
Successful target recovery does not necessarily imply
that other model behaviors remain unchanged. Internal
interventions may also affect outputs on inputs outside
the target-control set. To measure these effects, define
an off-target preservation set
\[
\Doff=\{\tilde x_j\}_{j=1}^{n_{\mathrm{off}}},
\]
containing $n_{\mathrm{off}}>0$ prompts that are
disjoint from the target inputs:
\[
\Doff\cap\{x_i:(x_i,\tau_i)\in\Dtar\}=\varnothing.
\]

For each input $x$, the unpatched model produces
an observable
\[
o_{\theta}(x)=o(f_{\theta},x),
\]
which is compared with the corresponding observable
$o(f_{\theta}^{\beta},x)$ after intervention.
The discrepancy $d_{\mathrm{off}}$ measures this change
and satisfies $d_{\mathrm{off}}(z,z)=0$ and
$d_{\mathrm{off}}(z,z')\geq0$.

For next-token probability distributions,
$d_{\mathrm{off}}$ may be the KL divergence from the
unpatched to the patched distribution. Other observables
may use squared Euclidean distance or task-specific
discrepancies. The preservation set, observable, and
discrepancy measure are specified as part of the control
problem, independently of the driver-selection algorithm.

\paragraph{Candidate control node library.}

Selecting an intervention site alone does not specify
how its activation should be modified. Each candidate
control node therefore specifies a physical site,
an intervention direction, and a selection cost:
\[
v=(s_v,u_v,c_v),
\]
where $s_v\in\N$ is the physical site,
$u_v(x)\in\R^{d_{s_v}}$ is the intervention direction
for input $x$, and $c_v\geq1$ is the selection cost.
The direction determines the activation perturbation
up to a scalar strength $\beta_v$.

These nodes form a finite candidate library
\[
\V=\{v_1,\ldots,v_M\},
\]
which defines the search space for driver selection.
Multiple nodes may share a physical site while using
different directions.

Directions are constructed from unpatched activations.
For a prescribed source--target pair $(x_i^r,x_i^c)$,
an input-dependent direction is
\[
u_v(x_i^r)
=h_{s_v}(x_i^c)-h_{s_v}(x_i^r).
\]
Alternatively, a direction shared across prompts can
be obtained from fitting-group means:
\[
u_v=\bar h_{s_v}^{(+)}-\bar h_{s_v}^{(-)},
\]
where $\bar h_{s_v}^{(+)}$ and $\bar h_{s_v}^{(-)}$
are the mean activations over the prescribed positive
and negative fitting groups, respectively.

Each direction rule is defined on all prompts in
the evaluation set $\Xeval$, with any required paired
inputs specified as part of the rule. The evaluation
set and direction rules remain fixed during optimization:
$u_v(x)$ may vary across inputs but does not depend
on the optimized coefficients $\beta$.

Head--position libraries contain nodes at specified
attention heads and token positions. Module-level
libraries use attention, MLP, or residual-stream
activations, while feature-based libraries use
sparse-autoencoder features or other learned coordinates.
Given $\V$, Driver-OMP and Coverage-Driver select
a compact support $S\subseteq\V$ and jointly refit
the intervention strengths on that support.

\paragraph{Inference-time patching and weighted sparsity.}
An intervention is represented by a coefficient vector
\[
\beta=(\beta_{v_1},\ldots,\beta_{v_M})\in\R^M,
\]
where $\beta_v=m_v\alpha_v$, with $m_v\in\{0,1\}$
indicating whether candidate control node $v$ is
selected and $\alpha_v\in\R$ specifying its strength.
During a forward pass on input $x$, the activation
at each physical site is modified additively:
\begin{equation}
h_s(x)\leftarrow h_s(x)+
\sum_{v:s_v=s}\beta_v u_v(x),
\qquad s\in\N.
\label{eq:rewrite_patch}
\end{equation}
The resulting patched model is denoted $f_\theta^\beta$.
Interventions modify activations during inference
without updating the model parameters $\theta$,
and $\beta=0$ recovers the original model.
Signed reversal in Section~\ref{sec:protocol}
applies $-\beta$ to the target-side input.

The active driver support is defined as
\[
S_\beta=\operatorname{supp}(\beta)
=\{v\in\V:\beta_v\ne0\}.
\]
Elements of $S_\beta$ are termed driver nodes.
For a selected support $S\subseteq\V$, $\beta_S$
denotes its coefficients, with all coefficients
outside $S$ set to zero. Thus $S_\beta\subseteq S$,
with equality when all selected coefficients are
nonzero.

The total selection cost is measured by
weighted sparsity:
\begin{equation}
\|\beta\|_{0,c}
=\sum_{v\in S_\beta}c_v.
\label{eq:rewrite_weighted_sparsity}
\end{equation}
For unit costs $c_v=1$, this reduces to
$|S_\beta|$. More generally, costs can account
for differences among intervention types, such
as attention-head outputs, residual activations,
and individual features. When multiple candidate
nodes share a physical site, weighted sparsity
counts their selection costs separately.

\paragraph{Intervention energy.}
Weighted sparsity measures the cost of selecting
driver nodes but does not bound the magnitude of
their activation perturbations. Define the
evaluation set
\[
\Xeval=\{x_i:(x_i,\tau_i)\in\Dtar\}\cup\Doff.
\]
With direction rules fixed on $\Xeval$, the
intervention energy is the mean squared magnitude
of the additive activation perturbations:
\begin{align}
E(\beta)
&=\frac{1}{|\Xeval|}
\sum_{x\in\Xeval}\sum_{s\in\N}
\left\|\sum_{v:s_v=s}\beta_v u_v(x)\right\|_2^2
=\beta^\top W\beta,
\label{eq:rewrite_energy}\\
W_{vw}
&=\begin{cases}
\displaystyle\frac{1}{|\Xeval|}
\sum_{x\in\Xeval}
\langle u_v(x),u_w(x)\rangle,
& s_v=s_w,\\[6pt]
0, & s_v\ne s_w.
\end{cases}\notag
\end{align}

The matrix $W\in\R^{M\times M}$ is symmetric
positive semidefinite. For a support $S\subseteq\V$,
the energy reduces to
$E(\beta)=\beta_S^\top W_{SS}\beta_S$ when
coefficients outside $S$ are zero.
Cross terms arise only between candidate
directions sharing the same physical site.

If each active node occupies a distinct site
and its direction has unit mean squared norm
over $\Xeval$, then
$W_{S_\beta S_\beta}=I$ and
$E(\beta)=\|\beta\|_2^2$.
Otherwise, $W$ accounts for the combined
activation perturbations at shared sites.

\paragraph{Target displacement induced by a patch.}
For an intervention $\beta$, the resulting
target-behavior displacement is
\begin{equation}
F(\beta)=
\big[b(f_\theta^\beta,x_i)
-b(f_\theta,x_i)\big]_{i=1}^{n}
\in\R^{nr}.
\label{eq:rewrite_F_beta}
\end{equation}
The target error and off-target loss are
\begin{align}
L_{\mathrm{tar}}(\beta)
&=\frac{1}{n}\sum_{i=1}^{n}
\|b(f_\theta^\beta,x_i)-\tau_i\|_2^2,
\label{eq:rewrite_Ltar}\\
L_{\mathrm{off}}(\beta)
&=\frac{1}{n_{\mathrm{off}}}
\sum_{j=1}^{n_{\mathrm{off}}}
d_{\mathrm{off}}\!\left(
o(f_\theta,\tilde x_j),
o(f_\theta^\beta,\tilde x_j)
\right).
\label{eq:rewrite_Loff}
\end{align}
By the definition of the target displacement $q$,
\[
L_{\mathrm{tar}}(\beta)
=\frac{1}{n}\|q-F(\beta)\|_2^2.
\]
These quantities are evaluated using the original
nonlinear Transformer and define the target-recovery
and preservation components of the control certificate.

\begin{definition}[Nonlinear control certificate;
restatement of Definition~\ref{def:certificate}]
\label{def:rewrite_certificate}
Fix a problem instance
$(f_\theta,\Dtar,\Doff,b,o,d_{\mathrm{off}},\V)$,
nonnegative tolerances $\varepsilon,\delta,\kappa$,
and a stabilizer $\xi>0$.
An intervention $\beta\in\R^M$ is an
$(\varepsilon,\delta,\kappa)$-control certificate
on the specified evaluation sets if all three
conditions hold:
\begin{equation}
\frac{\|q-F(\beta)\|_2}{\|q\|_2+\xi}
\leq\varepsilon,
\qquad
L_{\mathrm{off}}(\beta)\leq\delta,
\qquad
E(\beta)\leq\kappa.
\label{eq:rewrite_certificate}
\end{equation}
The tolerances bound normalized target error,
off-target change, and intervention energy,
respectively.
\end{definition}

Using
$L_{\mathrm{tar}}(\beta)=\|q-F(\beta)\|_2^2/n$,
the target-recovery condition is equivalently
\[
L_{\mathrm{tar}}(\beta)
\leq
\frac{\varepsilon^2(\|q\|_2+\xi)^2}{n}.
\]

\begin{definition}[Sparse internal control problem]
\label{def:rewrite_sic_problem}
For the fixed problem instance above, candidate
costs $\{c_v\}$, and an energy weight $\lambda_E\geq0$,
sparse internal control is formulated as
\begin{equation}
\min_{\beta\in\R^M}
\;\|\beta\|_{0,c}+\lambda_E E(\beta)
\quad
\text{s.t.}\quad
\beta\text{ satisfies Eq.~\eqref{eq:rewrite_certificate}}.
\label{eq:rewrite_sic_objective}
\end{equation}
The objective balances driver-selection cost
and intervention energy among interventions
satisfying the prescribed control tolerances.
\end{definition}

\paragraph{Budgeted formulation.}
For a prescribed selection-cost budget $k\geq0$,
a related formulation minimizes target and
off-target losses:
\begin{equation}
\begin{aligned}
\min_{\beta\in\R^M}\quad&
L_{\mathrm{tar}}(\beta)
+\mu L_{\mathrm{off}}(\beta)
+\lambda_E E(\beta)\\
\text{s.t.}\quad&
\|\beta\|_{0,c}\leq k,
\qquad E(\beta)\leq\kappa,
\end{aligned}
\label{eq:rewrite_budgeted_form}
\end{equation}
where $\mu\geq0$ weights off-target change.
Under unit costs, the selection-cost constraint
becomes $|S_\beta|\leq k$.

Equation~\eqref{eq:rewrite_sic_objective}
enforces target-recovery and preservation
tolerances as hard constraints.
Equation~\eqref{eq:rewrite_budgeted_form}
instead trades off these losses under fixed
selection-cost and energy budgets.
The formulations generally have different
feasible sets and minimizers.

Support selection is combinatorial, and the
behavioral losses are generally nonconvex in
$\beta$. The local proxy in Eq.~\eqref{eq:proxy}
guides support selection and strength refitting.
The resulting intervention is evaluated against
Eq.~\eqref{eq:rewrite_certificate} in the original
Transformer. Task-specific implementations are
specified in Appendix~\ref{app:hyper}.

\paragraph{Target specification and direction rules.}
Targets $\tau_i$ specify desired values in
behavioral-readout space, whereas $u_v(x)$
specifies a perturbation direction in activation
space. For paired reconstruction,
$x_i=x_i^r$ and $\tau_i=b(f_\theta,x_i^c)$.
A shared per-input target can instead be specified
by $\tau_i=\tau$ for all $i$; the required
displacements $q_i=\tau-b(f_\theta,x_i)$ may
still differ across inputs.
The feedback experiments in
Section~\ref{sec:protocol} prescribe targets
for the batch-mean readout and evaluate the
corresponding tracking error.

Direction construction is a separate choice.
Evaluation-pair directions (P) use activation
contrasts from prescribed input pairs, while
training-mean directions (F) use contrasts
averaged over the fitting split.
Either construction can be combined with
example-specific or shared per-input targets,
provided the required inputs are specified.

Within each evaluation setting, direction rules
are fixed on all target and preservation prompts.
The same rules determine patch application,
the energy matrix $W$, and certificate evaluation.
Task-specific direction constructions and
evaluation settings are given in
Appendices~\ref{app:tasks} and~\ref{app:hyper}.

\section{Theoretical analysis}
\label{sec:theory_rewrite}
\label{app:theory}

This section analyzes the local approximation used for sparse internal control. It characterizes the target-effect dictionary, the first-order reconstruction error of a fixed support, and the energy required for exact local interpolation. It then derives the energy-constrained ridge refit and proves monotone submodularity of the saturated coverage objective, with the standard greedy approximation bound under a pure cardinality constraint.

\subsection{Local effect dictionaries}
\label{subsec:local_effect_dictionaries}

For a fixed candidate library $\V$, define the nonlinear target displacement map
\[
\Phi(\beta)=F(\beta)=\big[b(f_{\theta}^{\beta},x_i)-b(f_{\theta},x_i)\big]_{i=1}^{n}\in\R^{nr}.
\]
The notation $\Phi=F$ is used for the target-displacement map in the proofs below. The candidate library, evaluation sets, and direction rules are fixed while the coefficients vary. For $S\subseteq\V$, $\Phi_S(\beta_S)$ denotes $F(\beta)$ with coefficients outside $S$ set to zero, and $G_S$ is the submatrix formed by the columns indexed by $S$. All Jacobians are evaluated at the unpatched model, $\beta=0$. In particular,
\[
J_{i,s_v}\in\R^{r\times d_{s_v}},\qquad
g_{i,v}\in\R^r,\qquad D\Phi_S(0)=G_S.
\]
Since $\Phi(0)=0$, the first-order approximation at the unpatched model is
\[
\Phi(\beta)\approx G\beta .
\]
The column associated with candidate $v$ is obtained by differentiating the behavioral readout with respect to the activation at the patched site:
\begin{equation}
 g_{i,v}=J_{i,s_v}u_v(x_i),
 \qquad
 J_{i,s_v}=\frac{\partial b(f_{\theta},x_i)}{\partial h_{s_v}(x_i)} .
\label{eq:rewrite_target_column}
\end{equation}
Stacking $g_{i,v}$ over target prompts gives $g_v\in\R^{nr}$, and stacking columns over candidates gives
\begin{equation}
G=[g_{v_1},\ldots,g_{v_M}]\in\R^{nr\times M}.
\label{eq:rewrite_G_dictionary}
\end{equation}
The local target equation is therefore
\begin{equation}
q\approx G\beta .
\label{eq:rewrite_local_target_eq}
\end{equation}

A corresponding off-target dictionary is built from a specified differentiable preservation surrogate $\psi_{\theta}(x)=\psi(f_\theta,x)\in\R^p$, where $p$ is its readout dimension. This surrogate may be the full next-token logit vector, a restricted or projected logit readout, or another differentiable observable. For each preservation prompt $\tilde x_j$ and candidate $v$, define
\begin{equation}
 h_{j,v}=K_{j,s_v}u_v(\tilde x_j),
 \qquad
 K_{j,s_v}=\frac{\partial \psi_{\theta}(\tilde x_j)}{\partial h_{s_v}(\tilde x_j)} .
\label{eq:rewrite_pres_column}
\end{equation}
Here $K_{j,s_v}\in\R^{p\times d_{s_v}}$ and $h_{j,v}\in\R^p$. Stacking over preservation prompts gives $h_v=[h_{1,v};\ldots;h_{n_{\mathrm{off}},v}]$ and $H=[h_{v_1},\ldots,h_{v_M}]\in\R^{n_{\mathrm{off}}p\times M}$. The metric is $\Omega=\operatorname{diag}(\Omega_1,\ldots,\Omega_{n_{\mathrm{off}}})$, where each $\Omega_j\in\R^{p\times p}$ is positive semidefinite; $\Omega^{1/2}$ denotes its symmetric positive-semidefinite square root.

For a squared-Euclidean discrepancy in the coordinates of $\psi$, $\Omega_j=I_p$. If $\psi$ is the full next-token logit vector and $p_j^0$ is the unpatched probability vector, its KL quadratic term is
\[
\frac12\,\Delta\psi_j^\top
\left[\operatorname{diag}(p_j^0)-p_j^0(p_j^0)^\top\right]
\Delta\psi_j,
\]
where $\Delta\psi_j$ is the logit displacement at $\tilde x_j$. The constant factor can be absorbed into $\mu$. For a restricted or projected readout, $\Omega_j$ specifies a surrogate metric in that readout's coordinates.

The sample-normalized matrices are
\[
\bar q=\frac{q}{\sqrt n},\qquad
\bar G=\frac{G}{\sqrt n},\qquad
\bar H=\frac{H}{\sqrt{n_{\mathrm{off}}}} .
\]
The local sparse proxy is
\begin{equation}
\min_{\beta\in\R^M}
\|\bar q-\bar G\beta\|_2^2
+\mu\|\Omega^{1/2}\bar H\beta\|_2^2
+\lambda\beta^{\top}W\beta
\quad
\text{s.t.}\quad
\|\beta\|_{0,c}\le k,
\;\beta^{\top}W\beta\le\kappa .
\label{eq:rewrite_local_proxy}
\end{equation}
Here $\mu\geq0$ weights predicted off-target change, $\lambda\geq0$ weights intervention energy in the local refit, and $k$ and $\kappa$ are the selection-cost and energy budgets. The local parameter $\lambda$ is distinct from $\lambda_E$ in the nonlinear objective. Equation~\eqref{eq:rewrite_local_proxy} combines first-order target reconstruction, a quadratic preservation surrogate, and energy regularization. The selected intervention is subsequently evaluated in the nonlinear model.

\subsection{Local linearization error}
\label{subsec:linearization_error_rewrite}

The next theorem states the basic locality guarantee.  It is stated for a fixed support $S\subseteq\V$, because the Hessian bound and the conditioning of the local dictionary depend on the selected sites.

\begin{theorem}[Local linearization error]
\label{thm:rewrite_local_linearization}
\label{thm:linearization-app}
Let $S\subseteq\V$, let $W_{SS}\succ0$, and let $\kappa\geq0$. Define the energy ball
\[
\mathcal B_{S,\kappa}=\{\beta_S\in\R^{|S|}:\beta_S^{\top}W_{SS}\beta_S\le\kappa\}.
\]
Assume $\Phi_S$ is twice continuously differentiable on an open neighborhood of $\mathcal B_{S,\kappa}$ and that its second derivative is bounded with a finite constant $L_S\geq0$: for all $\tilde\beta\in\mathcal B_{S,\kappa}$ and all $a,b\in\R^{|S|}$,
\begin{equation}
\left\|D^2\Phi_S(\tilde\beta)[a,b]\right\|_2
\le
L_S\|a\|_{W_{SS}}\|b\|_{W_{SS}},
\qquad
\|a\|_{W_{SS}}^2=a^{\top}W_{SS}a .
\label{eq:rewrite_hessian_bound}
\end{equation}
Here $D\Phi_S$ is the Jacobian, $D^2\Phi_S(\tilde\beta)[a,b]$ is the bilinear second derivative, and $\|a\|_{W_{SS}}=(a^\top W_{SS}a)^{1/2}$.
Then every $\beta_S\in\mathcal B_{S,\kappa}$ satisfies
\begin{equation}
\|\Phi_S(\beta_S)-G_S\beta_S\|_2
\le
\frac{L_S}{2}\,\beta_S^{\top}W_{SS}\beta_S .
\label{eq:rewrite_lin_error_bound}
\end{equation}
Equivalently, for the normalized displacement $\bar\Phi_S=\Phi_S/\sqrt n$ and $\bar G_S=G_S/\sqrt n$,
\begin{equation}
\|\bar\Phi_S(\beta_S)-\bar G_S\beta_S\|_2
\le
\frac{L_S}{2\sqrt n}\,\beta_S^{\top}W_{SS}\beta_S .
\label{eq:rewrite_norm_lin_error_bound}
\end{equation}
\end{theorem}

\begin{proof}
Since $\Phi_S(0)=0$ and $D\Phi_S(0)=G_S$, Taylor's theorem with integral remainder gives
\[
\Phi_S(\beta_S)-G_S\beta_S
=
\int_0^1(1-t)D^2\Phi_S(t\beta_S)[\beta_S,\beta_S]dt .
\]
Taking norms and applying Eq.~\eqref{eq:rewrite_hessian_bound} yields
\[
\|\Phi_S(\beta_S)-G_S\beta_S\|_2
\le
\int_0^1(1-t)L_S\|\beta_S\|_{W_{SS}}^2dt
=
\frac{L_S}{2}\beta_S^{\top}W_{SS}\beta_S .
\]
Dividing by $\sqrt n$ gives Eq.~\eqref{eq:rewrite_norm_lin_error_bound}.
\end{proof}

\begin{corollary}[Linear proposal to nonlinear target residual]
\label{cor:rewrite_proxy_to_nonlinear}
Under the assumptions of Theorem~\ref{thm:rewrite_local_linearization}, if $\beta_S\in\mathcal B_{S,\kappa}$, then
\begin{equation}
\frac{\|q-\Phi_S(\beta_S)\|_2}{\|q\|_2+\xi}
\le
\frac{\|q-G_S\beta_S\|_2}{\|q\|_2+\xi}
+
\frac{L_S\kappa/2}{\|q\|_2+\xi}.
\label{eq:rewrite_nonlinear_residual_bound}
\end{equation}
\end{corollary}

In particular, under the corollary's assumptions, the target-recovery inequality in Definition~\ref{def:certificate} holds if
\[
\|q-G_S\beta_S\|_2+L_S\kappa/2
\leq\varepsilon(\|q\|_2+\xi).
\]
The bound combines the local reconstruction error with a curvature contribution on the specified energy ball. Off-target preservation is evaluated through $L_{\rm off}$.

\subsection{Local target controllability of a support}
\label{subsec:local_controllability_rewrite}

For a fixed support $S$, the first-order question is whether the desired displacement $\bar q$ lies near the span of $\bar G_S$ under the energy budget.  Define the local controllability residual
\begin{equation}
\chi_{\kappa}(S)
=
\min_{\beta_S:\,\beta_S^{\top}W_{SS}\beta_S\le\kappa}
\frac{\|\bar q-\bar G_S\beta_S\|_2}{\|\bar q\|_2+\xi} .
\label{eq:rewrite_chi_kappa}
\end{equation}
A small $\chi_{\kappa}(S)$ indicates that the selected support can
approximate the target displacement within the energy budget at
first order. For any $\beta_S$ in the energy ball, Corollary~\ref{cor:rewrite_proxy_to_nonlinear}
bounds its nonlinear target residual through the local reconstruction
error and curvature term. Off-target preservation is assessed separately,
and Definition~\ref{def:certificate} combines target recovery, preservation,
and energy into the joint nonlinear certificate.

For $W_{SS}\succ0$, define the whitened dictionary
\[
\widetilde G_S=\bar G_S W_{SS}^{-1/2}.
\]
Here $W_{SS}^{-1/2}$ denotes the symmetric inverse square root, and $\widetilde G_S^{+}$ is the Moore--Penrose pseudoinverse. For $\bar q=0$, the minimum-energy interpolation is $\beta_S=0$. For nonzero $\bar q\in\operatorname{range}(\bar G_S)$, the whitened dictionary has at least one positive singular value.

\begin{theorem}[Minimum-energy local interpolation]
\label{thm:rewrite_min_energy_interpolation}
\label{thm:reachability-app}
Assume $W_{SS}\succ0$ and $0\ne\bar q\in\operatorname{range}(\bar G_S)$. Let $\sigma_{\min}^{+}(\widetilde G_S)$ be the smallest nonzero singular value of $\widetilde G_S$.  The minimum-energy solution of $\bar G_S\beta_S=\bar q$ is
\begin{equation}
\beta_S^{\dagger}=W_{SS}^{-1/2}\widetilde G_S^{+}\bar q,
\label{eq:rewrite_min_energy_solution}
\end{equation}
and it satisfies
\begin{equation}
(\beta_S^{\dagger})^{\top}W_{SS}\beta_S^{\dagger}
\le
\frac{\|\bar q\|_2^2}{\left(\sigma_{\min}^{+}(\widetilde G_S)\right)^2} .
\label{eq:rewrite_energy_singular_bound}
\end{equation}
\end{theorem}

\begin{proof}
Use whitened coordinates $z=W_{SS}^{1/2}\beta_S$.  The interpolation constraint becomes $\widetilde G_S z=\bar q$, and the energy becomes $\|z\|_2^2$.  The minimum-norm solution is $z^{\dagger}=\widetilde G_S^{+}\bar q$, whose norm is bounded by $\|\bar q\|_2/\sigma_{\min}^{+}(\widetilde G_S)$.  Substituting $\beta_S^{\dagger}=W_{SS}^{-1/2}z^{\dagger}$ proves the claim.
\end{proof}

For an attainable target $\bar q\in\operatorname{range}(\bar G_S)$, including $\bar q=0$, the exact minimum interpolation energy is
\[
E_{\min}(S)=\|\widetilde G_S^{+}\bar q\|_2^2.
\]
For $W_{SS}\succ0$, exact first-order reconstruction within budget is equivalent to
\[
\bar q\in\operatorname{range}(\bar G_S),\qquad
E_{\min}(S)\leq\kappa.
\]
Equation~\eqref{eq:rewrite_energy_singular_bound} supplies an upper bound on this energy for a nonzero attainable target; the target's alignment with the singular directions determines its exact value.

The normalization in Eq.~\eqref{eq:rewrite_chi_kappa} differs from that in Corollary~\ref{cor:rewrite_proxy_to_nonlinear}. For a minimizer $\beta_S^\chi$ in Eq.~\eqref{eq:rewrite_chi_kappa},
\[
\|q-G_S\beta_S^\chi\|_2
=(\|q\|_2+\sqrt n\,\xi)\chi_\kappa(S).
\]
Under Theorem~\ref{thm:rewrite_local_linearization}, Corollary~\ref{cor:rewrite_proxy_to_nonlinear} therefore gives
\[
\frac{\|q-\Phi_S(\beta_S^\chi)\|_2}{\|q\|_2+\xi}
\leq
\frac{\|q\|_2+\sqrt n\,\xi}{\|q\|_2+\xi}\chi_\kappa(S)
+\frac{L_S\kappa}{2(\|q\|_2+\xi)}.
\]
The shared refit in Eq.~\eqref{eq:rewrite_refit_problem} also includes preservation and ridge terms and does not generally minimize $\chi_\kappa(S)$. Its coefficients are assessed using their own local residual in Corollary~\ref{cor:rewrite_proxy_to_nonlinear}.

\subsection{Energy-constrained ridge refit}
\label{subsec:ridge_refit_theory_rewrite}\label{app:kkt-refit}

After a support $S$ is selected, strengths are recomputed by a shared refit.  Define the augmented target and dictionary
\begin{equation}
\hat q=\begin{bmatrix}\bar q\\0\end{bmatrix},
\qquad
\hat G_S=\begin{bmatrix}
\bar G_S\\
\sqrt\mu\,\Omega^{1/2}\bar H_S
\end{bmatrix} .
\label{eq:rewrite_augmented_dictionary}
\end{equation}
The refit solves
\begin{equation}
\min_{\beta_S\in\R^{|S|}}
\|\hat q-\hat G_S\beta_S\|_2^2+\lambda\beta_S^{\top}W_{SS}\beta_S
\quad
\text{s.t.}\quad
\beta_S^{\top}W_{SS}\beta_S\le\kappa .
\label{eq:rewrite_refit_problem}
\end{equation}
When $W_{SS}\succ0$ and $\kappa=0$, the unique feasible coefficient vector is $\beta_S=0$. The following multiplier characterization concerns $\kappa>0$.

\begin{proposition}[KKT form of the shared refit]
\label{prop:rewrite_refit_kkt}
Assume $W_{SS}\succ0$, $\lambda>0$, and $\kappa>0$. The unique minimizer of Eq.~\eqref{eq:rewrite_refit_problem} is
\begin{equation}
\beta_S^{\star}(\nu)=
\left(\hat G_S^{\top}\hat G_S+(\lambda+\nu)W_{SS}\right)^{-1}\hat G_S^{\top}\hat q,
\label{eq:rewrite_kkt_solution}
\end{equation}
where $\nu\geq0$ satisfies primal feasibility, $(\beta_S^{\star})^{\top}W_{SS}\beta_S^{\star}\leq\kappa$, and complementary slackness:
\begin{equation}
\nu\left((\beta_S^{\star})^{\top}W_{SS}\beta_S^{\star}-\kappa\right)=0 .
\label{eq:rewrite_kkt_complementarity}
\end{equation}
If $E(\beta_S^{\star}(0))\leq\kappa$, then $\nu=0$. Otherwise, a unique positive multiplier satisfies $E(\beta_S^{\star}(\nu))=\kappa$.
\end{proposition}

\begin{proof}
The objective is strictly convex because $\lambda W_{SS}\succ0$, and $\beta_S=0$ is strictly feasible because $\kappa>0$. Thus the KKT conditions are necessary and sufficient. With
\[
\mathcal L(\beta_S,\nu)=
\|\hat q-\hat G_S\beta_S\|_2^2+
\lambda\beta_S^{\top}W_{SS}\beta_S+
\nu(\beta_S^{\top}W_{SS}\beta_S-\kappa),
\]
stationarity gives
\[
A_\nu\beta_S=\hat G_S^\top\hat q,
\qquad
A_\nu=\hat G_S^\top\hat G_S+(\lambda+\nu)W_{SS}\succ0.
\]
This proves Eq.~\eqref{eq:rewrite_kkt_solution}. Along the multiplier path,
\[
\frac{d}{d\nu}E(\beta_S^{\star}(\nu))
=-2(W_{SS}\beta_S^{\star})^\top A_\nu^{-1}
(W_{SS}\beta_S^{\star})\leq0.
\]
The energy is continuous, tends to zero as $\nu\to\infty$, and is strictly decreasing when $\beta_S^{\star}\ne0$. If the unconstrained energy exceeds $\kappa>0$, a unique positive boundary multiplier therefore exists and can be found by bisection.
\end{proof}

\subsection{Coverage objective and submodularity}
\label{subsec:coverage_submodularity_rewrite}

Residual-fitting objectives can over-allocate support to rows that are already easy to explain.  For population-level steering, a useful driver should move many prompts in the correct direction, even if no single prompt has a large residual.  \CoverageDriver\ therefore uses a saturated row-coverage score.

Here $j\in\{1,\ldots,nr\}$ indexes a coordinate of the stacked target readout. For $r=1$, one row corresponds to one prompt; for $r>1$, it corresponds to a prompt--readout-coordinate pair. Let $\V^{\pm}$ denote the candidate library augmented with both $u_v$ and $-u_v$. In Eqs.~\eqref{eq:rewrite_coverage_a}--\eqref{eq:rewrite_coverage_objective}, the dictionary columns and candidate sets refer to this oriented library. The convention $\operatorname{sign}(0)=0$ is used. For each row $j$ and candidate $v$, define a nonnegative signed contribution
\begin{equation}
a_{jv}=\operatorname{clip}\!\left(
\frac{\operatorname{sign}(\bar q_j)\bar G_{jv}}{|\bar q_j|+\delta_0},0,1
\right),
\label{eq:rewrite_coverage_a}
\end{equation}
where $\delta_0>0$ prevents division by very small target entries. The orientation expansion represents a direction useful with a negative strength as a positive-oriented candidate during coverage scoring. Given row weights $w_j\ge0$, define
\begin{equation}
C(S)=\sum_j w_j\min\!\left(1,\sum_{v\in S}a_{jv}\right).
\label{eq:rewrite_coverage_objective}
\end{equation}
The cap at $1$ implements diminishing returns: once a row is covered, additional candidates targeting the same row provide no further coverage reward.

\begin{proposition}[Monotone submodularity of saturated coverage]
\label{prop:rewrite_coverage_submodular}
If $a_{jv}\ge0$ and $w_j\ge0$, then $C(S)$ in Eq.~\eqref{eq:rewrite_coverage_objective} is monotone submodular and $C(\varnothing)=0$. For an integer $1\leq k\leq|\V^{\pm}|$, let $S^{\star}\in\arg\max_{S\subseteq\V^{\pm},\,|S|\leq k}C(S)$. Under this pure cardinality budget with no additional candidate-exclusion constraints, the standard $k$-step greedy algorithm that repeatedly adds the largest marginal increase obtains
\[
C(S_{\mathrm{greedy}})\ge(1-1/e)C(S^{\star}).
\]
\end{proposition}

\begin{proof}
For each row $j$, the set function $m_j(S)=\sum_{v\in S}a_{jv}$ is modular and nondecreasing.  The scalar function $\phi(t)=\min(1,t)$ is concave and nondecreasing on $\R_{\ge0}$.  The composition $\phi(m_j(S))$ is therefore monotone submodular.  A nonnegative weighted sum of monotone submodular functions is monotone submodular.  The cardinality-constrained greedy guarantee is the classical result of \citet{nemhauser1978}.
\end{proof}

For heterogeneous costs, define $\Delta_v(S)=C(S\cup\{v\})-C(S)$. The score $\Delta_v(S)/(c_v+\eta_0)$, with $\eta_0>0$, provides a cost-normalized greedy rule. Proposition~\ref{prop:rewrite_coverage_submodular} concerns the pure cardinality-constrained coverage objective. Executed selection and stopping settings are specified in Appendix~\ref{app:hyper}.

\paragraph{Scope of the guarantee.}
The \((1-1/e)\) bound concerns the nonnegative saturated coverage
objective in Proposition~\ref{prop:rewrite_coverage_submodular} under
a pure cardinality constraint and standard \(k\)-step greedy selection.
It does not automatically apply to additional site or orientation
exclusions, early-stopped variants, or heterogeneous-cost greedy rules.
Nor does it guarantee nonlinear target movement after strength refitting.
The formal \(\pm u_v\) expansion supplies candidate orientations for
support proposal; the final \textsc{SharedRefit} may assign either sign
to a coefficient. The executed MCQA and Refusal Coverage variants do not
expand both orientations (Appendix~\ref{app:hyper}).

\subsection{Relation to orthogonal matching pursuit}
\label{subsec:omp_relation_rewrite}
\label{thm:omp-app}

\DriverOMP\ is a structured variant of orthogonal matching pursuit.  With unit-normalized effect-dictionary columns, no off-target block, $W=I$, $c_v=1$, $\lambda=0$, unconstrained refitting, and no additional candidate exclusions, the residual-fitting selection rule reduces to ordinary OMP.  Classical OMP recovery theorems apply when the dictionary satisfies strong incoherence or restricted-isometry assumptions \citep{tropp2007signal}.  Transformer effect dictionaries are typically redundant and highly correlated, so the algorithm is used as a sparse proposal method rather than as an exact-support recovery procedure.  A refitted intervention receives the nonlinear control certificate when all three conditions in Definition~\ref{def:certificate} are satisfied.

\subsection{End-to-end interpretation}
\label{subsec:end_to_end_theory_rewrite}

The analysis characterizes first-order target error, the minimum energy for local interpolation, and the energy-constrained fixed-support refit. Saturated coverage supplies a complementary objective for support proposal. \DriverOMP\ and \CoverageDriver\ use residual fitting and row coverage, respectively, to construct candidate supports. Evaluation in the patched nonlinear model measures target recovery, off-target preservation, and intervention energy against Definition~\ref{def:certificate}.

\section{Driver-selection algorithms}
\label{sec:algorithms_rewrite}
\label{app:algorithms}

This section specifies the formal support-selection, energy-constrained refitting, and nonlinear-validation procedures. \DriverOMP\ uses residual fitting, whereas \CoverageDriver\ uses saturated row coverage. Each selector returns a support and refitted coefficients; \textsc{ValidateCertificate} then reports the three certificate metrics and an acceptance decision. The executed experimental configurations are specified in Appendix~\ref{app:hyper}.

\subsection{Shared inputs and outputs}
\label{subsec:algorithm_inputs_rewrite}

The shared inputs include the frozen model $f_\theta$, target and preservation sets $\Dtar,\Doff$, behavioral readout $b$, preservation observable $o$ and divergence $d_{\rm off}$, differentiable surrogate $\psi$, preservation metric $\Omega$, candidate library $\V$, fixed direction rules, costs $c_v\geq1$, and prescribed candidate-exclusion and tie-breaking rules. Here $k_c\geq0$ denotes the selection-cost budget written as $k$ in the weighted-sparsity constraints of Sections~\ref{sec:problem}--\ref{sec:algorithms}; under unit costs it bounds the number of selected candidates. The formal refit uses $\mu\geq0$, $\lambda>0$, and $\kappa\geq0$. The library and admissibility rules are assumed to give $W_{SS}\succ0$ for every nonempty support passed to the positive-definite refit of Appendix~\ref{subsec:ridge_refit_theory_rewrite}. Site-disjoint candidates with nonzero mean squared direction norms satisfy this condition. The ordinary-OMP specialization in Appendix~\ref{subsec:omp_relation_rewrite} uses unconstrained least-squares refitting.

A selected set $S$ and its coefficient vector $\beta_S$ define a full-library vector by setting coefficients outside $S$ to zero. Its active support satisfies $S_\beta\subseteq S$. Support selection enforces the cost budget and specified exclusions; \textsc{SharedRefit} enforces the energy constraint. Algorithms~\ref{alg:rewrite_driver_omp}--\ref{alg:rewrite_coverage_driver} return proposals. Algorithm~\ref{alg:rewrite_validate} accepts a proposal when all three conditions of Definition~\ref{def:certificate} are satisfied.

The numerical parameters are a target-normalization stabilizer $\xi>0$, a cost stabilizer $\eta_0>0$, a coverage stabilizer $\delta_0>0$, a relative linear-residual tolerance $\varepsilon_{\rm lin}\in[0,1)$, an improvement threshold $\eta_{\rm tol}\geq0$, a coverage threshold $\tau\in(0,1]$, and a normalized gain threshold $g_{\min}\geq0$. \textsc{SharedRefit} uses a multiplier-interval tolerance $t_\nu>0$ and a relative energy-deficit tolerance $t_E\in(0,1)$; these are numerical tolerances, not the certificate thresholds $(\varepsilon,\delta,\kappa)$.

\subsection{Building local dictionaries}
\label{subsec:build_tables_rewrite}

The target dictionary $G$ stores the first-order effect of every candidate on every target row, and $H$ stores its first-order effect on the preservation surrogate. The Gram matrix $W$ represents the energy of simultaneous patches. The evaluation set is $\Xeval=\{x_i:(x_i,\tau_i)\in\Dtar\}\cup\Doff$, with $n,n_{\rm off}>0$. Each direction rule is defined at every input in this set and remains fixed while the intervention coefficients vary. All derivatives are evaluated at the unpatched model, $\beta=0$. The returned objects are $(\bar q,\bar G,\bar H,W,\Omega)$. For the energy matrix, inner products are evaluated only between directions at the same physical site; entries for different sites are set to zero, as in Appendix~\ref{sec:sic_problem_rewrite}.

\begin{algorithm}[!htbp]
\small
\caption{\textsc{BuildLocalTables}: target, preservation, and energy dictionaries}
\label{alg:rewrite_build_tables}
\begin{algorithmic}[1]
\Require Frozen model $f_\theta$; sets $\Dtar,\Doff$ with $n,n_{\rm off}>0$; library $\V$ with fixed site/direction rules; readouts $b,\psi$; metric $\Omega$.
\Ensure Normalized target $\bar q$, dictionaries $\bar G,\bar H$, energy matrix $W$, and metric $\Omega$.
\State $\Xeval\gets\{x_i:(x_i,\tau_i)\in\Dtar\}\cup\Doff$.
\State Evaluate $q_i\gets\tau_i-b(f_\theta,x_i)$, stack $q$, and set $\bar q\gets q/\sqrt n$.
\For{each $v\in\V$}
  \For{each target input $x_i$}
    \State $g_{i,v}\gets J_{i,s_v}u_v(x_i)$, with $J_{i,s_v}$ evaluated at $\beta=0$.
  \EndFor
  \For{each preservation input $\tilde x_j$}
    \State $h_{j,v}\gets K_{j,s_v}u_v(\tilde x_j)$, with $K_{j,s_v}$ evaluated at $\beta=0$.
  \EndFor
\EndFor
\State Stack columns to form $G,H$; set $\bar G\gets G/\sqrt n$, $\bar H\gets H/\sqrt{n_{\rm off}}$.
\For{each pair $v,w\in\V$}
  \If{$s_v=s_w$}
    \State $W_{vw}\gets |\Xeval|^{-1}\sum_{x\in\Xeval}\langle u_v(x),u_w(x)\rangle$.
  \Else
    \State $W_{vw}\gets0$.
  \EndIf
\EndFor
\State \Return $(\bar q,\bar G,\bar H,W,\Omega)$.
\end{algorithmic}
\end{algorithm}

In practice, the products $J_{i,s_v}u_v(x_i)$ and $K_{j,s_v}u_v(\tilde x_j)$ can be computed by automatic differentiation as Jacobian--vector products or vector--Jacobian products, depending on whether the readout dimension is small or large. For scalar margins, a single backward pass per example can score many candidate directions at a site. For vector-valued panels, rows are accumulated across readout coordinates.

\subsection{Shared energy-constrained refit}
\label{subsec:shared_refit_algorithm_rewrite}

On a fixed support, \textsc{SharedRefit} jointly solves the quadratic refit in Appendix~\ref{subsec:ridge_refit_theory_rewrite}. Let
\[
\hat q=\begin{bmatrix}\bar q\\0\end{bmatrix},
\qquad
\hat G_S=\begin{bmatrix}\bar G_S\\\sqrt\mu\,\Omega^{1/2}\bar H_S\end{bmatrix}.
\]
The refit solves
\[
\min_{\beta_S}
\|\hat q-\hat G_S\beta_S\|_2^2+
\lambda\beta_S^{\top}W_{SS}\beta_S
\quad
\text{s.t.}\quad
\beta_S^{\top}W_{SS}\beta_S\le\kappa .
\]
It returns an empty vector for an empty support and zero coefficients when $\kappa=0$. For a nonempty support with $\kappa>0$, it first tests the unconstrained ridge solution. If the energy budget is active, it brackets the multiplier and bisects while retaining a feasible upper endpoint. The Gram matrix and right-hand side are computed once for each support and reused in the linear solves. Bisection stops when either the multiplier interval is at most $t_\nu$ or the feasible endpoint's energy deficit is at most $t_E\kappa$. Formal reference values are $t_\nu=10^{-8}$ and $t_E=10^{-3}$ (Appendix~\ref{app:hyper}).

\begin{algorithm}[!htbp]
\small
\caption{\textsc{SharedRefit}: energy-constrained ridge refit}
\label{alg:rewrite_shared_refit}
\begin{algorithmic}[1]
\Require $S$, $\hat q$, $\hat G_S$, $W_{SS}\succ0$ for nonempty $S$; $\lambda>0$; $\kappa\geq0$; $t_\nu>0$, $0<t_E<1$.
\Ensure Coefficients on $S$ with energy at most $\kappa$.
\If{$S=\varnothing$}
  \State \Return the empty coefficient vector.
\EndIf
\If{$\kappa=0$}
  \State \Return $0\in\R^{|S|}$.
\EndIf
\State $B_S\gets\hat G_S^\top\hat G_S$; $g_S\gets\hat G_S^\top\hat q$.
\Function{Solve}{$\nu$}
  \State Solve $[B_S+(\lambda+\nu)W_{SS}]\beta=g_S$.
  \State \Return $(\beta,\beta^\top W_{SS}\beta)$.
\EndFunction
\State $(\beta,e)\gets\Call{Solve}{0}$.
\If{$e\leq\kappa$}
  \State \Return $\beta$.
\EndIf
\State $\nu_{\rm lo}\gets0$; $\nu_{\rm hi}\gets1$; $(\beta_{\rm hi},e_{\rm hi})\gets\Call{Solve}{\nu_{\rm hi}}$.
\While{$e_{\rm hi}>\kappa$}
  \State $\nu_{\rm lo}\gets\nu_{\rm hi}$; $\nu_{\rm hi}\gets2\nu_{\rm hi}$.
  \State $(\beta_{\rm hi},e_{\rm hi})\gets\Call{Solve}{\nu_{\rm hi}}$.
\EndWhile
\While{$\nu_{\rm hi}-\nu_{\rm lo}>t_\nu$ \textbf{and} $\kappa-e_{\rm hi}>t_E\kappa$}
  \State $\nu\gets(\nu_{\rm lo}+\nu_{\rm hi})/2$; $(\beta,e)\gets\Call{Solve}{\nu}$.
  \If{$e>\kappa$}
    \State $\nu_{\rm lo}\gets\nu$.
  \Else
    \State $(\nu_{\rm hi},\beta_{\rm hi},e_{\rm hi})\gets(\nu,\beta,e)$.
  \EndIf
\EndWhile
\State \Return $\beta_{\rm hi}$.
\end{algorithmic}
\end{algorithm}

\subsection{\DriverOMP: sparse residual fitting}
\label{subsec:driver_omp_algorithm_rewrite}

\DriverOMP\ ranks candidates using their alignment with the remaining augmented residual. Define the augmented full-library column for candidate $v$ as
\[
\hat g_v=\begin{bmatrix}\bar G_v\\\sqrt\mu\,\Omega^{1/2}\bar H_v\end{bmatrix}.
\]
At iteration $t$, with residual $r_t=\hat q-\hat G_{S_t}\beta_{S_t}$, the candidate-ranking score is

\begin{equation}
\operatorname{score}(v\mid S_t)=
\frac{|\langle \hat g_v,r_t\rangle|^2}
{(\|\hat g_v\|_2^2+\lambda W_{vv})(c_v+\eta_0)} .
\label{eq:rewrite_omp_score}
\end{equation}

Equation~\eqref{eq:rewrite_omp_score} is a cost-normalized, regularized correlation score for candidate ranking. The augmented columns include the linear preservation surrogate, and the selected coefficients are jointly recomputed by \textsc{SharedRefit}. The stopping statistic is the relative linear target residual. A trial support is retained when it reaches $\varepsilon_{\rm lin}$; otherwise, it must reduce that residual by more than $\eta_{\rm tol}$.

\begin{algorithm}[!htbp]
\small
\caption{\DriverOMP: residual-fitting driver selection}
\label{alg:rewrite_driver_omp}
\begin{algorithmic}[1]
\Require $\bar q,\bar G,\bar H,W,\Omega$, library $\V$, costs $c_v$, exclusions, budget $k_c\geq0$; $\lambda>0$, $\mu\geq0$, $\kappa\geq0$; $\xi,\eta_0>0$, $0\leq\varepsilon_{\rm lin}<1$, $\eta_{\rm tol}\geq0$; $t_\nu,t_E$.
\Ensure Proposal $(S,\beta_S)$; nonlinear validation is a separate step.
\State Form $\hat q\gets[\bar q;0]$ and $\hat G\gets[\bar G;\sqrt\mu\,\Omega^{1/2}\bar H]$.
\State $S\gets\varnothing$; $\beta_S\gets[\,]$; $r\gets\hat q$.
\State $\rho_{\rm prev}\gets\|\bar q\|_2/(\|\bar q\|_2+\xi)$.
\If{$\rho_{\rm prev}\leq\varepsilon_{\rm lin}$ \textbf{or} $\kappa=0$}
  \State \Return $(S,\beta_S)$.
\EndIf
\While{$\sum_{v\in S}c_v<k_c$}
  \State $\mathcal A\gets\{v\in\V\setminus S:\sum_{u\in S}c_u+c_v\leq k_c\text{ and exclusions hold}\}$.
  \If{$\mathcal A=\varnothing$}
    \State \textbf{break}.
  \EndIf
  \State Compute $\operatorname{score}(v\mid S)$ from Eq.~\eqref{eq:rewrite_omp_score} for every $v\in\mathcal A$.
  \State Select a maximizer $v^\star$ using the prescribed tie rule.
  \State $S'\gets S\cup\{v^\star\}$.
  \State $\beta'\gets\Call{SharedRefit}{S',\hat q,\hat G_{S'},W_{S'S'},\lambda,\kappa,t_\nu,t_E}$.
  \State $\rho'\gets\|\bar q-\bar G_{S'}\beta'\|_2/(\|\bar q\|_2+\xi)$.
  \If{$\rho'\leq\varepsilon_{\rm lin}$}
    \State $(S,\beta_S)\gets(S',\beta')$; \textbf{break}.
  \EndIf
  \If{$\rho_{\rm prev}-\rho'\leq\eta_{\rm tol}$}
    \State \textbf{break}. \Comment{Return the previous support; reject the trial.}
  \EndIf
  \State $(S,\beta_S)\gets(S',\beta')$; $r\gets\hat q-\hat G_S\beta_S$; $\rho_{\rm prev}\gets\rho'$.
\EndWhile
\State \Return $(S,\beta_S)$.
\end{algorithmic}
\end{algorithm}

\subsection{\CoverageDriver: saturated row coverage}
\label{subsec:coverage_driver_algorithm_rewrite}

\CoverageDriver\ selects candidates using a saturated target-row coverage objective. Let $\V^{\pm}$ be the oriented candidate library and write $v^\sigma=(s_v,\sigma u_v,c_v)$ for $\sigma\in\{-1,+1\}$. All matrices passed to the oriented selector use the same indexing:
\[
\bar G_{v^\sigma}=\sigma\bar G_v,\qquad
\bar H_{v^\sigma}=\sigma\bar H_v,\qquad
W_{v^\sigma,w^{\sigma'}}=\sigma\sigma'W_{vw}.
\]
The costs and physical-site identifiers are inherited from the original candidates. After fitting oriented coefficients, the equivalent original coefficient is $\beta_v=\beta_{v^+}-\beta_{v^-}$, with an absent orientation contributing zero. Selection applies all specified site and orientation exclusions.

The index $j\in\{1,\ldots,nr\}$ refers to a target-readout coordinate. For $r=1$ each row corresponds to a prompt. The row weights satisfy $w_j\geq0$, and $W_{\rm row}=\sum_jw_j$ is a scalar weight sum, distinct from the energy matrix $W$. If $W_{\rm row}=0$, the coverage objective is identically zero and the selector returns the empty support. The convention $\operatorname{sign}(0)=0$ is used. For each row and oriented candidate, compute
\[
a_{jv}=\operatorname{clip}\!\left(
\frac{\operatorname{sign}(\bar q_j)\bar G_{jv}}{|\bar q_j|+\delta_0},0,1
\right).
\]
Let $\chi_j(S)=\min(1,\sum_{v\in S}a_{jv})$ be the current coverage of row $j$. The marginal gain of adding $v$ is

\begin{equation}
\Delta(v\mid S)=\sum_j w_j\left[\min(1,\chi_j(S)+a_{jv})-\chi_j(S)\right].
\label{eq:rewrite_coverage_gain}
\end{equation}

The selected candidate maximizes $\Delta(v\mid S)/(c_v+\eta_0)$ among feasible candidates.

\begin{algorithm}[!htbp]
\small
\caption{\CoverageDriver: saturated row-coverage driver selection}
\label{alg:rewrite_coverage_driver}
\begin{algorithmic}[1]
\Require $\bar q$; oriented library $\V^\pm$ and consistently oriented $\bar G,\bar H,W$; $\Omega$; $w_j\geq0$; $c_v$, exclusions, $k_c\geq0$; $\tau\in(0,1]$, $g_{\min}\geq0$, $\delta_0,\eta_0>0$; $\lambda>0$, $\mu\geq0$, $\kappa\geq0$, $t_\nu,t_E$.
\Ensure Oriented proposal $(S,\beta_S)$; map coefficients back before original-library validation.
\State $S\gets\varnothing$; $\chi_j\gets0$ for all $j$; $W_{\rm row}\gets\sum_jw_j$.
\If{$W_{\rm row}=0$ \textbf{or} $\kappa=0$}
  \State \Return $(\varnothing,[\,])$.
\EndIf
\State $a_{jv}\gets\operatorname{clip}(\operatorname{sign}(\bar q_j)\bar G_{jv}/(|\bar q_j|+\delta_0),0,1)$, with $\operatorname{sign}(0)=0$.
\While{$\sum_{v\in S}c_v<k_c$}
  \State $\mathrm{cov}\gets W_{\rm row}^{-1}\sum_jw_j\chi_j$.
  \If{$\mathrm{cov}\geq\tau$}
    \State \textbf{break}.
  \EndIf
  \State $\mathcal A\gets\{v\in\V^\pm\setminus S:\sum_{u\in S}c_u+c_v\leq k_c\text{ and exclusions hold}\}$.
  \If{$\mathcal A=\varnothing$}
    \State \textbf{break}.
  \EndIf
  \State Compute $\Delta(v\mid S)$ from Eq.~\eqref{eq:rewrite_coverage_gain} for every $v\in\mathcal A$.
  \State Select $v^\star\in\arg\max_{v\in\mathcal A}\Delta(v\mid S)/(c_v+\eta_0)$ using the prescribed tie rule.
  \If{$\Delta(v^\star\mid S)\leq g_{\min}W_{\rm row}$}
    \State \textbf{break}.
  \EndIf
  \State $S\gets S\cup\{v^\star\}$; $\chi_j\gets\min(1,\chi_j+a_{jv^\star})$ for every $j$.
\EndWhile
\State Form $\hat q\gets[\bar q;0]$, $\hat G\gets[\bar G;\sqrt\mu\,\Omega^{1/2}\bar H]$.
\State $\beta_S\gets\Call{SharedRefit}{S,\hat q,\hat G_S,W_{SS},\lambda,\kappa,t_\nu,t_E}$.
\State \Return $(S,\beta_S)$.
\end{algorithmic}
\end{algorithm}

Saturation assigns marginal credit to target rows that remain uncovered. \CoverageDriver\ therefore allocates its support through row coverage, whereas \DriverOMP\ ranks candidates using the current residual. The coverage score defines the support-proposal objective; joint intervention strengths are obtained in the subsequent refit.

\subsection{Nonlinear validation}
\label{subsec:nonlinear_validation_algorithm_rewrite}

The output of either selector is a proposal. Validation applies the additive patch in the original Transformer and recomputes the nonlinear quantities in the certificate. For the candidate library and evaluation sets supplied to validation, $R_{\rm tar}=\mathrm{Rel}_{\rm tar}(1)$ denotes the target residual in Definition~\ref{def:certificate}. The same direction rules define the patched model and its energy matrix.

\begin{algorithm}[!htbp]
\small
\caption{\textsc{ValidateCertificate}: nonlinear acceptance test}
\label{alg:rewrite_validate}
\begin{algorithmic}[1]
\Require $f_\theta$; original candidate library with site/direction rules; $(S,\beta_S)$; $\Dtar=\{(x_i,\tau_i)\}$ with $\tau_i\in\R^r$; $\Doff$; readouts $b,o$; divergence $d_{\rm off}$; energy matrix $W$ for these rules and sets; $\xi>0$, tolerances $(\varepsilon,\delta,\kappa)$.
\Ensure Acceptance status and $(R_{\rm tar},L_{\rm off}(\beta),E(\beta))$.
\State If necessary, map oriented coefficients to the original library as specified in Section~\ref{subsec:coverage_driver_algorithm_rewrite}.
\State Form $\beta$ by zero extension of the original-library coefficients.
\State Compute $q_i\gets\tau_i-b(f_\theta,x_i)$ and stack $q$.
\State Evaluate the patched model and set $F(\beta)\gets[b(f_\theta^\beta,x_i)-b(f_\theta,x_i)]_{i=1}^{n}$.
\State $R_{\rm tar}\gets\|q-F(\beta)\|_2/(\|q\|_2+\xi)=\mathrm{Rel}_{\rm tar}(1)$.
\State $L_{\rm off}(\beta)\gets n_{\rm off}^{-1}\sum_jd_{\rm off}(o(f_\theta,\tilde x_j),o(f_\theta^\beta,\tilde x_j))$.
\State $E(\beta)\gets\beta^\top W\beta$.
\If{$R_{\rm tar}\leq\varepsilon$ \textbf{and} $L_{\rm off}(\beta)\leq\delta$ \textbf{and} $E(\beta)\leq\kappa$}
  \State \Return Accept, $(R_{\rm tar},L_{\rm off}(\beta),E(\beta))$.
\Else
  \State \Return Reject, $(R_{\rm tar},L_{\rm off}(\beta),E(\beta))$.
\EndIf
\end{algorithmic}
\end{algorithm}

Validation returns \textsc{Accept} or \textsc{Reject} together with $R_{\rm tar}$, $L_{\rm off}(\beta)$, and $E(\beta)$ for the specified intervention and evaluation sets. The selection-cost and exclusion constraints are enforced by the support-selection stage.

\subsection{Selection recurrence and consensus drivers}
\label{subsec:consensus_algorithm_rewrite}

Sparse dictionaries in Transformers can be redundant: multiple supports may produce similar target movement. Selection frequency summarizes a specified collection of configuration runs. For the reported IOI analysis, five seeds and five support budgets $k\in\{3,5,8,12,16\}$ give $B=25$ configurations. The sets $S^{(b)}$ are the final supports returned by the IOI recurrence implementation, which includes target-readout evaluation, fallback selection, and pruning. In that implementation, the validation flag uses mean absolute error to the clean target, and the stored scores also include $R_{\rm clean}$. These run-specific checks are distinct from \textsc{ValidateCertificate} in Algorithm~\ref{alg:rewrite_validate}. The configurations are not bootstrap resamples. For candidate $v$, the frequency is
\[
\pi(v)=\frac{1}{B}\sum_{b=1}^{B}\mathbf 1[v\in S^{(b)}].
\]
For a threshold $\pi^{\star}$, the consensus core is
\[
C_{\pi^{\star}}=\{v:\pi(v)\ge\pi^{\star}\}.
\]
The frequency $\pi(v)$ measures recurrence across this configuration grid, and $\pi^\star$ defines the consensus set studied in Appendix~\ref{app:rq1}. The consensus core is not used to replace the nonlinear certificate. It is a descriptive recurrence diagnostic: candidates that appear across the specified configurations can be studied further, but the frequency alone is not a selection probability, an uncertainty interval, or a guarantee on unseen prompts.

\subsection{Computational complexity}
\label{subsec:algorithm_complexity_rewrite}

Let $d=nr+n_{\rm off}p$ be the number of augmented rows, let $M=|\V|$, and let $K$ bound the number of candidate additions permitted by the budget. For a nonempty candidate library, set $c_{\min}=\min_v c_v$; then
\[
K\leq\min(M_{\rm sel},\lfloor k_c/c_{\min}\rfloor),
\]
where $M_{\rm sel}$ is the number of columns scored by the selector, including any orientation expansion. For a fixed support of size $s$, precomputing $\hat G_S^\top\hat G_S$ and $\hat G_S^\top\hat q$ costs $O(ds^2)$. If the refit uses $T_s$ dense linear solves, its cost is $O(ds^2+T_s s^3)$; $T_s$ includes the initial solve, multiplier bracketing, and bisection. For a bracket of width $B_\nu$, bisection to an absolute multiplier tolerance $t_\nu$ takes at most
\[
\max\{0,\lceil\log_2(B_\nu/t_\nu)\rceil\}
\]
steps; the energy-deficit condition can terminate it earlier.

For $K\geq1$, with $T=\max_{1\leq s\leq K}T_s$, \DriverOMP\ has the dictionary-level bound
\[
O(KMd+dK^3+TK^4),
\]
whereas \CoverageDriver\ has the bound
\[
O(KM_{\pm}nr+dK^2+TK^3),
\]
with $M_{\pm}=|\V^{\pm}|$. These expressions assume constant-time candidate-exclusion checks and an incrementally maintained support cost, and count work after the augmented dictionaries are constructed. Dictionary differentiation, energy-matrix construction, nonlinear validation, and any experimental fallback or pruning calls are accounted for separately. \DriverOMP\ refits each trial support, including a final rejected trial if present; \CoverageDriver\ performs one final refit.

\subsection{Algorithmic summary}
\label{subsec:algorithm_summary_rewrite}

\textsc{BuildLocalTables} constructs the target, preservation, and energy quantities. \DriverOMP\ or \CoverageDriver\ returns a candidate support with energy-constrained coefficients, and \textsc{ValidateCertificate} evaluates that intervention in the nonlinear model. The complete procedure returns the intervention, its three certificate metrics, and the acceptance status. Selection recurrence summarizes the supports of a separately specified collection of runs. Executed experimental variants and their recorded checks are described in Appendices~\ref{app:hyper}--\ref{app:empirical}.

\section{Control evaluation protocol: complete definitions}
\label{app:protocol-app}

This section gives the complete definitions of the six control axes
introduced in Section~\ref{sec:protocol}: reachability, signed reversal,
dose response, feedback controllability, off-target preservation,
and held-out reuse. Experimental operating criteria are specified in
Appendix~\ref{app:hyper}, and configurations are indexed in
Table~\ref{tab:experiment-configurations}.

An evaluated intervention specifies a support $S$, base coefficients
$\beta_S$, and a direction rule $u_v(x)$. A dose $\gamma\in\R$
defines the patched model $f_\theta^{\gamma\beta}$, with $\gamma=1$
denoting the nominal dose. Target displacement is
$F(\gamma\beta)=[b(f_\theta^{\gamma\beta},x_i)-b(f_\theta,x_i)]_{i=1}^n$.
The six axes report properties of this intervention in the nonlinear
model. Definition~\ref{def:certificate} separately specifies the joint
target-recovery, off-target-preservation, and intervention-energy conditions.

\subsection{Reachability}
\label{app:protocol-reachability}

Reachability is measured by the relative residual between the desired
displacement $q$ and the displacement actually produced by the patch:
\begin{equation}
\mathrm{Rel}_{\mathrm{tar}}(\gamma)=
\frac{\|q-F(\gamma\beta)\|_2}{\|q\|_2+\xi} .
\label{eq:reltar-app}
\end{equation}
The stabilizer $\xi>0$ regularizes normalization near zero requested
displacement. At nominal dose, $\mathrm{Rel}_{\rm tar}(1)$ measures
the normalized remaining target displacement and is compared with
the specified tolerance $\varepsilon$. For scalar readouts, an auxiliary
target-sign agreement rate compares the signs of the patched and
target readouts. Here $b_{\rm scalar}=b$ and
$\tau_{i,\rm scalar}=\tau_i$ when $r=1$:
\begin{equation}
\operatorname{SignAgree}(\beta)=\frac{1}{n}\sum_{i=1}^{n}
\mathbf 1\!\bigl[\operatorname{sign}(b_{\mathrm{scalar}}(f_\theta^\beta,x_i))=
\operatorname{sign}(\tau_{i,\mathrm{scalar}})\bigr].
\label{eq:flip-app}
\end{equation}
Agreement is measured at nominal dose, with $\operatorname{sign}(0)=0$.
For IOI, Sentiment, and Refusal, $b_{\rm scalar}$ is the specified
task readout. The repeated MCQA controls use the
correct-minus-mean-wrong margin $M_{\rm MCQA}$ defined in
Section~\ref{sec:exp_setup}, with the reference answer index fixed
across each pair. Nominal MCQA reconstruction uses the four option
logits as a vector; it is a separate configuration and does not replace
that scalar definition. Nominal reconstruction additionally reports
$R_{\rm clean}$, which normalizes by the clean-output norm
(Appendix~\ref{app:reachability-results}).

\subsection{Signed reversal}
\label{app:protocol-reversal}

Signed reversal applies $-\beta$ to the target-side member of a
prescribed pair while retaining the selected support and the original
direction convention. Write $x_i^r$ for the source member and $x_i^c$
for the target member. In the paired-contrast condition, the original
target-minus-source contrast is retained when the negative patch is
applied. The subscript $\mathrm{dis}$ denotes the resulting target-side
readout after that patch:
\begin{align}
\mathrm{Rel}_{\mathrm{dis}\to\mathrm{src}}
&=\frac{\bigl\|[b(f_\theta^{-\beta},x_i^c)-b(f_\theta,x_i^r)]_{i=1}^{n}\bigr\|_2}{D},\nonumber\\
\mathrm{Rel}_{\mathrm{dis}\to\mathrm{clean}}
&=\frac{\bigl\|[b(f_\theta^{-\beta},x_i^c)-b(f_\theta,x_i^c)]_{i=1}^{n}\bigr\|_2}{D},\nonumber\\
D&=\bigl\|[b(f_\theta,x_i^c)-b(f_\theta,x_i^r)]_{i=1}^{n}\bigr\|_2+\xi,\nonumber\\
\mathrm{Reversal}(\beta)
&=\frac{\mathrm{Rel}_{\mathrm{dis}\to\mathrm{clean}}}
{\mathrm{Rel}_{\mathrm{dis}\to\mathrm{src}}}.
\label{eq:reversal-app}
\end{align}
A ratio greater than one places the reverse-patched readout closer
to the source readout than to the unpatched target readout.
If $\mathrm{Rel}_{\mathrm{dis}\to\mathrm{src}}=0$ and
$\mathrm{Rel}_{\mathrm{dis}\to\mathrm{clean}}>0$, the ratio is
$+\infty$; if both distances vanish, it is undefined.
These cases are recorded explicitly. The common normalizer $D$
cancels from the ratio. Operating criteria and per-seed results are
given in Appendices~\ref{app:hyper} and~\ref{app:formal-reversal-results}.

\subsection{Dose response}
\label{app:protocol-dose}

Dose response measures changes in target progress across a specified
dose grid $\Gamma$, with support, base coefficients, and the direction
rule fixed. Negative doses reverse the patch sign on the source input;
$\gamma=0$ is the unpatched model and $\gamma=1$ the nominal dose.
This source-side scan is distinct from the target-side reversal test.
The grid is specified for each experimental series. For vector readouts,
projected target progress is
\begin{equation}
P(\gamma)=\frac{\langle q,F(\gamma\beta)\rangle}{\|q\|_2^2+\xi}.
\label{eq:progress-app}
\end{equation}
The signed Spearman coefficient $\rho_{\rm sp}$, denoted $\rho$
in the result figures and tables, summarizes the ordering of dose
and target progress without requiring a linear response.
Scalar mean-margin scans use $P_{\rm mean}$ from
Section~\ref{sec:protocol}, distinct from the vector projection in
Eq.~\eqref{eq:progress-app}. The sign of the coefficient records
whether progress increases or decreases with dose. Fixed-grid scans
and controller-trajectory diagnostics are identified separately.
Task-specific grids and operating criteria are reported in
Appendices~\ref{app:hyper} and~\ref{app:dose-results}.

\subsection{Feedback controllability}
\label{app:protocol-feedback}

Feedback controllability is evaluated by tracking specified scalar
targets with dose updates based on observed error. The selected
support, base coefficients, and direction rule remain fixed.
For an evaluation batch, the mean readout is
\begin{equation}
M(\gamma)=\frac{1}{n}\sum_i b_{\mathrm{scalar}}(f_\theta^{\gamma\beta},x_i)
\label{eq:mean-margin-app}
\end{equation}
and the tracking error is $e_t=m^\star-M(\gamma_t)$, following
Section~\ref{sec:protocol}. A reference proportional controller is
\begin{align}
e_t&=m^\star-M(\gamma_t),\nonumber\\
\gamma_{t+1}&=\operatorname{clip}\!\bigl(\gamma_t+\eta_t\,e_t,\gamma_{\min},\gamma_{\max}\bigr),\nonumber\\
\eta_{t+1}&=
\begin{cases}
\eta_t/2, & \operatorname{sign}(e_{t+1})\neq \operatorname{sign}(e_{t}),\\
\eta_t, & \text{otherwise}.
\end{cases}
\label{eq:controller-app}
\end{align}
Here $t$ indexes controller updates, $\eta_t>0$ is the gain, and
$[\gamma_{\min},\gamma_{\max}]$ is the dose interval of this
reference controller. The error $e_{t+1}$ is evaluated at the
updated dose before the next gain is selected.

Equation~\eqref{eq:controller-app} specifies a reference controller.
The executed MCQA and Refusal experiments use the fixed-gain update
and stopping rules in Appendix~\ref{app:hyper}, without dose clipping
or gain halving. A terminal hit satisfies
$|m^\star-M(\gamma_{\rm end})|\leq\tau_{\rm ctrl}$.
Endpoint errors, hit counts, and recorded readout-evaluation counts
are reported by seed and target; evaluation counts are distinct from
controller update counts. The recorded count includes the initial
zero-dose readout and final endpoint readout; zero-dose calls may reuse
the cached unpatched baseline. Initial zero-dose hits are identified separately.
Targets within one seed share the fitted intervention and evaluation batch.

\subsection{Off-target preservation}
\label{app:protocol-preservation}

Off-target preservation measures changes in the predictive distribution
on the specified preservation set $\Doff$. The nominal-dose loss is
\begin{equation}
L_{\mathrm{off}}(\beta)=\frac{1}{n_{\mathrm{off}}}\sum_{j=1}^{n_{\mathrm{off}}}
\mathrm{KL}\!\left(p_\theta(\cdot\mid \tilde x_j)\,\|\,p_\theta^\beta(\cdot\mid \tilde x_j)\right).
\label{eq:loff-app}
\end{equation}
Evaluation at dose $\gamma$ uses $L_{\rm off}(\gamma\beta)$.
The unpatched distribution is the first KL argument, and the patched
distribution is evaluated at the same readout position. KL divergence
is computed using natural logarithms.
For a specified tolerance $\delta$, the seed-mean rule compares
the average prompt KL with $\delta$, whereas the all-prompt rule
requires every prompt KL to meet $\delta$. The panel size,
operating threshold, and per-seed outcomes are given in
Appendices~\ref{app:hyper} and~\ref{app:preservation-results}.

The IOI paired-position diagnostic instead measures
$\mathrm{KL}(p_{\rm patched}\Vert p_{\rm clean})$ at non-answer
positions of paired prompts and reports the change relative to the
unpatched-corrupt baseline. The unpaired Sentiment dose plot uses
$\mathrm{KL}(p_{\rm unpatched}\Vert p_{\rm patched})$ on its
negative evaluation prompts. Appendix~\ref{app:legacy-dose} records
these panel-specific definitions.

The auxiliary ratio
\begin{equation}
\mathrm{Sel}(\gamma)=
\frac{\|F(\gamma\beta)\|_2/(\|q\|_2+\xi)}{\sqrt{L_{\mathrm{off}}(\gamma\beta)}+\xi}.
\label{eq:selectivity-app}
\end{equation}
divides normalized displacement magnitude by
$\sqrt{L_{\rm off}(\gamma\beta)}+\xi$.
Its numerator is a magnitude; $\mathrm{Rel}_{\rm tar}(\gamma)$
separately measures proximity to the requested target.
The six-axis preservation measure remains $L_{\rm off}(\gamma\beta)$.

\subsection{Held-out reuse}
\label{app:protocol-transfer}

Held-out reuse evaluates intervention components on prompts excluded
from support selection. Four information conditions are distinguished.

\paragraph{Refit-$\beta$ reuse.}
Selected sites remain fixed, while coefficients are refitted using
the held-out target readouts and prescribed evaluation-pair directions.

\paragraph{Fixed-$\beta$ reuse with evaluation-pair directions.}
Sites and training-fitted coefficients remain fixed, while directions
use activation contrasts from the evaluation pairs.

\paragraph{Feedback with training-mean directions (F).}
Sites, base coefficients, and training-mean directions remain fixed,
while dose is adjusted using evaluation-time feedback.
This condition is evaluated under feedback controllability.

\paragraph{Fully frozen open-loop reuse.}
Sites, coefficients, training-mean directions, and dose are all fixed
before evaluation. This is the complete-intervention condition
called frozen-$\beta$ reuse in Section~\ref{sec:protocol}.
P/F identifies the direction source; it does not by itself specify
whether the full intervention is frozen.

Each series retains its recorded split and statistic. Module-level
MCQA uses four seeds per category, 32 questions split 16/16, and
signed mean-margin recovery
$(\bar b_{\rm patched}-\bar b_{\rm source})/
(\bar b_{\rm target}-\bar b_{\rm source})$, with means over the
evaluation batch and absolute denominators below $10^{-9}$ omitted.
Sentiment's eight-seed paired fixed-dose producer uses the same
numerator and $\max(\bar b_{\rm target}-\bar b_{\rm source},10^{-6})$
in the denominator. Three-seed IOI reuse instead reports
one minus the mean pointwise clean-normalized absolute error, with
evaluation-pair directions retained in its fixed-coefficient branch.
These recovery statistics are distinct from
$1-\mathrm{Rel}_{\rm tar}^{\rm test}(1)$.
Both Sentiment conditions retain training-fitted coefficients:
P uses evaluation-pair directions and F uses training-mean directions.
Recorded recovery values are reported in Appendix~\ref{app:rq4};
series-specific statistical procedures are given in
Appendix~\ref{app:reproducibility}.

\section{Task instantiations}
\label{app:tasks}

This section specifies each task's model, prompt construction,
behavioral readout, paired targets, direction rules, and candidate
library. The notation follows Sections~\ref{sec:problem}--\ref{sec:protocol}:
$b$ is the behavioral readout, $\tau_i$ the target readout,
$u_v$ a candidate direction rule, $\V$ the candidate library,
and $\Doff$ the preservation panel. Fitting and evaluation samples
are indexed in Table~\ref{tab:experiment-configurations}; numerical
settings are given in Appendix~\ref{app:hyper}.

\paragraph{Preservation readouts.}
The differentiable surrogate $\psi$ determines the local dictionary
$H$; its coordinate metric $\Omega$ is specified with the surrogate
(Appendix~\ref{subsec:local_effect_dictionaries}). The nonlinear
observable for the independent preservation panels is the full
next-token distribution, $o(f_\theta,x)=p_\theta(\cdot\mid x)$, with
\[
d_{\rm off}(p_\theta,p_\theta^{\gamma\beta})
=\mathrm{KL}\!\left(p_\theta\;\|\;p_\theta^{\gamma\beta}\right).
\]
Baseline and patched forwards use the same specified input and readout
position, with the patch's direction rule defined for that input.
IOI paired-position and unpaired Sentiment diagnostics are defined in
Appendix~\ref{app:legacy-dose}. Executed refits and penalty terms are
specified in Appendix~\ref{app:hyper}.

\paragraph{Paired targets and feedback targets.}
For paired fitting and reconstruction, $\tau_i$ is the target member's
readout under the specified experimental configuration. On IOI this
is the clean counterpart of the corrupt source; MCQA and the paired
Sentiment/Refusal controls use their configuration-specific target
members. Feedback uses separately specified batch-mean targets
$m^\star$, whose executed grids are listed in Appendix~\ref{app:hyper}.
Those grids are not assignments to the paired-fitting $\tau_i$.

For a prescribed source--target pair $(x_i^r,x_i^c)$, the
evaluation-pair direction and fitting-split mean direction are
\[
u_v^{\rm P}(x_i^r)=h_{s_v}(x_i^c)-h_{s_v}(x_i^r),\qquad
u_v^{\rm F}=\frac{1}{n_{\rm fit}}\sum_{i\in\mathcal I_{\rm fit}}
\bigl[h_{s_v}(x_i^c)-h_{s_v}(x_i^r)\bigr],
\]
where $n_{\rm fit}=|\mathcal I_{\rm fit}|$.
P and F specify direction information; coefficient fitting and dose
updates are stated separately for each series. The unpaired Sentiment
dose series uses the class-mean construction described below.

\subsection{Indirect Object Identification}
\label{app:task-ioi}

The IOI experiments use GPT-2 small with twelve layers and twelve
attention heads per layer. Prompts follow paired ABBA/BABA templates
such as ``Then, $A$ and $B$ went to the place. $B$ gave an item to''.
Templates and vocabulary items are sampled using the recorded seed
and constructor. The behavioral readout is the indirect-object-minus-subject
logit margin at the answer position. The reference IO and subject token
identities are taken from the clean member and kept fixed when scoring
both members. The source is the corrupt member, and its target readout
is the clean member's margin.

END is the final valid prompt position; S1 and S2 are the first and
second subject occurrences, and IO is the indirect-object occurrence
in each member's role assignment. Position rules are resolved separately
for the two members. The library spans twelve layers, twelve heads,
and END, IO, S1, and S2, giving 576 candidates. The nominal reconstruction
uses 64 pairs and ten seeds, with the same panel used for fitting and
evaluation. The separate feedback, dose, reuse, and recurrence series
retain the configurations in Table~\ref{tab:experiment-configurations}.

The IOI feedback and dose producers use raw clean-minus-corrupt
activation contrasts at head--position sites. Candidate-score column
normalization is distinct from unit-normalizing activation differences.
The IOI paired-position diagnostic measures full-vocabulary KL
at non-answer positions of the same paired prompts: patched relative
to clean, with the unpatched corrupt-relative-to-clean KL reported
separately as a baseline. Baseline subtraction applies only to quantities
explicitly defined as $\Delta_{\rm off}$ or $\Delta$KL. This diagnostic
uses a different panel and KL reference from the independent
preservation measurements.

The additional IOI reversal keeps the ten-seed nominal supports and
coefficients, applies the opposite patch to the target member using
each evaluation pair's contrast, and uses the same 64-pair panel.
It adds no frozen-direction IOI condition.

\subsection{Multiple-choice question answering}
\label{app:task-mcqa}

\paragraph{Nominal reconstruction.}
The nominal reconstruction in Table~\ref{tab:nominal-restored} uses
Qwen3-4B \citep{qwen3} on the Humanities and STEM subject categories
of MMLU. The readout is the vector of four option logits.
The library contains \texttt{resid\_post}, \texttt{mlp\_out}, and
\texttt{attn\_out} at each of 36 layers, giving 108 module-level
candidates. Each category uses eight questions on the same fitting
and evaluation panel over three seeds. Table~\ref{tab:nominal-restored}
reports $\mathrm{Rel}_{\mathrm{tar}}(1)$; Table~\ref{tab:nominal-clean}
provides the clean-output-normalized reconstruction statistic
$R_{\rm clean}$ (Appendix~\ref{app:reachability-results}).

\paragraph{Head-level control.}
The repeated scalar-control series uses 1,152 answer-position
attention-head candidates: 36 layers times 32 heads. Each seed draws
24 category-filtered MMLU test questions and divides them into twelve
fitting and twelve evaluation examples. The source member exchanges
the correct option with a seeded incorrect option; the target member
retains the original order. The original answer index $a^\star$ is
used for both members in the correct-minus-mean-wrong readout
$M_{\rm MCQA}$ from Section~\ref{sec:exp_setup}. Supports and base
coefficients are fitted on the fitting half; P and F use evaluation-pair
and training-mean directions, respectively. Feedback, signed reversal,
and independent dose response use eight seeds per category.

\paragraph{Module-level held-out reuse.}
The separate reuse series uses the scalar correct-minus-mean-wrong
readout $M_{\rm MCQA}$ with the module-level library. It uses 32
questions per category, a 16/16 fitting/evaluation split, and four seeds.
Refit-$\beta$ retains the selected support with evaluation-side
coefficient fitting and paired directions. Fully frozen open-loop reuse
applies the fitting-side support, coefficients, mean directions, and
nominal dose. Its recovery statistic and fitting normalization are
specified in Appendices~\ref{app:protocol-transfer} and~\ref{app:hyper}.

The additional STEM dose scan reuses the saved head-level supports
and coefficients in both P and F conditions. The independent preservation
panel contains sixteen Humanities and sixteen STEM questions. Both
baseline and patched forwards use the original unswapped panel prompt;
its option-swapped companion supplies the clean-minus-swapped activation
contrast. Each prompt is processed without padding, with the patch and
full-vocabulary KL readout at the final valid input position. The native
support and coefficients remain fixed at nominal dose. Thus this panel
uses evaluation-pair directions, separately from fully frozen reuse.

\subsection{Sentiment steering}
\label{app:task-sentiment}

Sentiment uses GPT-2 small. Its positive list is \{good, great,
wonderful, amazing, happy, excellent, fantastic, lovely\}; the negative
list is \{bad, terrible, awful, horrible, miserable, disappointing,
boring, depressing\}. Each word is encoded with a leading space and
\texttt{add\_special\_tokens=False}; only its first subtoken is used.
The readout is the mean next-token log-probability of the positive
IDs minus that of the negative IDs, not a whole-word sequence probability.

The unpaired-dose producer constructs pools of sixteen
positive and sixteen negative templates, with eight fitting and eight
evaluation examples per class. Selection and directions use fitting-side
population means. Its progress divides the patched-minus-negative
evaluation mean by the positive-minus-negative mean gap, floored at
$10^{-6}$. Its KL uses the same negative evaluation prompts
(unpatched relative to patched), rather than an independent panel.
The fitting/evaluation split and producer settings are specified in
\path{configs/sentiment_unpaired_dose.json}, with the implementation
registered in \path{experiments/registry.json}.
The three-seed curves in Figure~\ref{fig:dose} summarize the saved
unpaired-dose records identified in the accompanying source index.

The paired fixed-dose comparison contains sixteen template pairs per
seed, shuffled into eight fitting and eight evaluation pairs. The negative
template is the source and the positive template is the target.
Five of the sixteen template pairs have identical source and target
strings and therefore zero requested displacement; these pairs are
retained in both fitting and evaluation according to the seeded split.
The library has 144 last-position attention-head candidates.
Support selection and ridge fitting use the fitting half.
At nominal dose $\gamma=1$, P uses evaluation-pair activation contrasts
and F uses fitting-split mean contrasts; both retain training-fitted
coefficients. Eight seeds and support budgets $k\in\{3,5,8\}$ are used
for this comparison. Its mean-margin recovery retains the denominator
specified in Appendix~\ref{app:protocol-transfer}.
Excluding the zero-displacement pairs from the saved evaluation
vectors leaves the $\varepsilon=0.4$ pass counts unchanged for all
three selectors and both direction conditions.

Additional target-side reversal and independent preservation reuse
only the saved $k=8$ interventions. Reversal tests both P and F on
the eight evaluation pairs per seed, applying $-\beta$ to the positive
target member. Preservation broadcasts fitting-split mean directions
on the independent 32-prompt MMLU panel at nominal dose.
Sentiment dose response and fixed-dose reuse are evaluated in
separate unpaired and paired experiments, respectively. The reported
Sentiment series contain no multi-target feedback experiment, so
feedback controllability remains unassessed for this task.

\subsection{Refusal steering}
\label{app:task-refusal}

The Refusal experiments use Qwen3-1.7B and Arditi-inspired
harmful/harmless instruction pairs \citep{arditi2024refusal}.
The refusal list is \{I, Sorry, Unfortunately, Cannot, Can\}; the
compliance list is \{Sure, Of, Here, Let, Certainly\}. Each word is
encoded with a leading space and \texttt{add\_special\_tokens=False},
retaining its first subtoken. The readout is the mean refusal-ID
minus mean compliance-ID next-token log-probability. This lexical
readout is evaluated before generating a token, at the input position
specified by the producer's ANSWER rule (Appendix~\ref{app:hyper}).
It measures a lexical margin rather than completed-answer semantic safety.

In the paired controls, the harmless instruction is the source $x_i^r$
and the harmful instruction is the target $x_i^c$. Fitting uses
harmful-minus-harmless activation contrasts. Forward control applies
$\beta$ to the harmless member; signed reversal applies $-\beta$ to
the harmful member using the same direction convention. Feedback
targets are specified separately from the paired target readouts.

The library spans 28 layers and sixteen heads at ANSWER, giving 448
candidates. Twenty instruction pairs are shuffled by seed and split
into ten fitting and ten evaluation pairs. P and F are both evaluated,
with support and base coefficients fitted on the fitting half.
The nominal-dose preservation test uses the independent 32-prompt
MMLU panel with fitting-split mean directions.

The separate short-generation probe evaluates ten harmful prompts per
seed across eight seeds under four conditions: the unpatched model
and the three selectors. It generates exactly sixteen greedy tokens
per continuation, giving 320 continuations on twenty prompt identities
reused across seeds. Decoding and patch-position rules are specified
in Appendix~\ref{app:hyper}; outcomes are reported in Appendix~\ref{app:rq6}.

\section{Hyperparameters and implementation details}
\label{app:hyper}

\paragraph{Refit objectives and normalization.}
The formal refit in Appendices~\ref{app:theory}--\ref{app:algorithms}
uses sample-normalized target and preservation terms with an
activation-energy penalty and constraint. The repeated scalar MCQA
and Refusal runs fit the unnormalized training dictionary by
coefficient ridge:
\[
\min_{\beta_S}\|q_{\rm fit}-G_{{\rm fit},S}\beta_S\|_2^2
+\lambda\|\beta_S\|_2^2,\qquad \lambda=1,
\]
with solution
\[
\widehat\beta_S=(G_{{\rm fit},S}^{\top}G_{{\rm fit},S}+\lambda I)^{-1}
G_{{\rm fit},S}^{\top}q_{\rm fit}.
\]
These runs use raw activation contrasts and omit the preservation block
and independent energy-ball constraint. The directions are not
unit-normalized, so coefficient ridge does not generally penalize the
activation-space energy $E(\beta)=\beta^\top W\beta$.
The conditions for equality with $\|\beta\|_2^2$ are stated in
Appendix~\ref{sec:sic_problem_rewrite}. For fixed directions,
$E(\gamma\beta)=\gamma^2E(\beta)$; its evaluation requires $W$ on
the specified evaluation set. The numerical stabilizer is $\xi=10^{-6}$.

Nominal module-level MCQA uses $\lambda=10^{-5}$, while nominal IOI
uses $\lambda=1$, with their raw flattened dictionaries and displacements.
The retained paired and unpaired Sentiment producers likewise fit raw
dictionaries; their recorded reconstruction recipes specify
\texttt{-{}-ridge 1}. Module-level MCQA reuse uses $\lambda=1$ after
dividing both its dictionary and displacement by the square root of the
flattened target-readout row count $n_{\rm row}$. The relative penalty
scale therefore differs:
\[
\frac{1}{n_{\rm row}}\|q-G\beta\|_2^2+\lambda\|\beta\|_2^2
\quad\Longleftrightarrow\quad
\|q-G\beta\|_2^2+n_{\rm row}\lambda\|\beta\|_2^2.
\]
Each series retains its own scaling and refit parameter.
The paired and unpaired Sentiment configurations are provided in
\path{configs/sentiment_paired.json} and
\path{configs/sentiment_unpaired_dose.json}.
Module-level MCQA reuse is configured in
\path{configs/mcqa_humanities_reuse.json} and
\path{configs/mcqa_stem_reuse.json}.
Their execution modules are registered in
\path{experiments/registry.json}.

\paragraph{Intervention coordinates.}
The retained TransformerLens head-position adapter patches the selected
head slice at \texttt{hook\_z}, before the attention output projection.
Its position rule selects the specified token position. The module-level
adapter defines \texttt{resid\_post} as the post-block residual,
\texttt{mlp\_out} as the MLP output, and \texttt{attn\_out} as the
attention-module output. Tokenizer and position rules remain specific
to each experimental series.

\paragraph{Executed support selection.}
For the repeated scalar MCQA and Refusal controls, the executed OMP score is
\[
\operatorname{score}_{\mathrm{exec}}(v\mid S_t)=
\frac{|\langle G_v,r_t\rangle|^2}
{\max(\|G_v\|_2^2,10^{-12})\,c_v},
\qquad c_v=1.
\]
There is no \(\lambda W_{vv}\) term in this selection denominator;
ridge enters the subsequent strength refit. The OMP stopping defaults
are residual tolerance \(10^{-3}\) and improvement threshold \(10^{-4}\).
The executed OMP implementation evaluates
$\|q-G_S\widehat\beta_S\|_2/\max(\|q\|_2,10^{-12})$
on the current support at the start of each iteration; a stagnation
exit retains that already selected support. This differs from the
trial acceptance and rollback rule in Algorithm~\ref{alg:rewrite_driver_omp}.
Coverage uses \texttt{signed\_split=False}, \texttt{cover\_mode=soft},
\texttt{weight\_mode=absq}, \texttt{eps\_norm}\(=10^{-6}\),
\texttt{coverage\_target}\(=0.95\), and
\texttt{min\_marginal\_gain}\(=10^{-4}\). Its executed row weights and
contributions are
\[
d_j=\max(|q_j|,10^{-6}),\qquad w_j=d_j,
\qquad a_{jv}=\operatorname{clip}\!\left(
\frac{\operatorname{sign}(q_j)G_{jv}}{d_j},0,1\right).
\]
The implementation thus uses a maximum floor, rather than the additive
floor of the formal contribution in Eq.~\eqref{eq:rewrite_coverage_a}.
Negative alignment is clipped during proposal; final ridge coefficients
may have either sign. Coverage is accumulated as
\(C(S)=\sum_jw_j\min(1,\sum_{v\in S}a_{jv})\), with stopping based on
\(C(S)/\sum_jw_j\) and marginal gain normalized by \(\sum_jw_j\).
The formal settings in Table~\ref{tab:hyper} do not override these
executed values.

\paragraph{Module-level MCQA reuse implementation.}
The module-level held-out reuse records use a separate selector
implementation and explicit \texttt{refit\_mode=ridge}. Its target
dictionary and displacement are normalized by the square root of the
flattened target-readout row count. The OMP defaults are residual
tolerance $0.05$, minimum marginal gain $10^{-3}$, and cost stabilizer
$\eta_0=10^{-4}$; these differ from the scalar-control defaults above.
The selected training and evaluation dictionaries are refitted by
plain ridge for their respective reuse conditions. The optional
constrained-refit and certificate helpers are not called for these
recorded runs.

\paragraph{Executed feedback and fixed-dose settings.}
For the repeated MCQA and Refusal controls, the support budget is at
most eight, ridge is \(\lambda=1\), and the three selectors are
\DriverOMP{}, \CoverageDriver{}, and Top-\(k\). Top-\(k\) ranks
\(\lvert G_{\rm train}^{\top}q_{\rm train}\rvert\); all three fit
the selected coefficients on the fitting split. Feedback observes the
evaluation batch's mean margin and updates
\(\gamma_{t+1}=\gamma_t+\eta(m^\star-M(\gamma_t))\), starting from
\(\gamma_0=0\), with no dose clipping or gain halving. The iteration
cap is twelve, followed by the producer's final endpoint evaluation.
MCQA uses \(\eta=0.15\), endpoint tolerance \(0.3\), and targets
\(\{-2,-1,0,1,2,3\}\) for both categories. Refusal uses
\(\eta=0.5\), tolerance \(0.1\), and targets
\(\{-1.55,-1.45,-1.35,-1.25,-1.15\}\). A hit is determined from
the final realized batch-mean error, including any target already
within tolerance before a nonzero intervention. These executed settings
are distinct from the clipped reference controller in
Appendix~\ref{app:protocol-feedback}.

The separate IOI feedback record contains three seeds and eight
batch-mean targets $\{-3,-1,0,1,2,3,4,5\}$, with endpoint tolerance
$\tau_{\rm ctrl}=0.3$ and the same 64-prompt fitting/evaluation panel.
Its executable settings are specified in
\path{configs/ioi_feedback.json}.

\paragraph{Dose-response grids and additional measurements.}
The independent Humanities, STEM, and Refusal scans use
\[
\Gamma=\{-1,-0.5,0,0.25,0.5,0.75,1,1.25,1.5,2\}.
\]
The IOI dose record uses the same ten points.
The unpaired Sentiment dose record uses
\[
\Gamma_{\rm Sentiment}=\{0,0.25,0.5,1,1.5,2,3\}.
\]
Support, base coefficients, and direction construction remain fixed
within each scan; the series retain their own sample sizes, seeds,
and support budgets. Additional IOI/Sentiment reversal applies target-side
$-\beta$ without reselection or refitting. The added Sentiment reversal
and preservation use the saved $k=8$ controls, rather than the full
$k\in\{3,5,8\}$ fixed-dose comparison. The Refusal preservation panel
uses training-mean directions at $\gamma=1$ on 32 prompts reused across
eight seeds. MCQA uses the panel-specific pairing and input member in
Appendix~\ref{app:task-mcqa}.

\paragraph{Refusal formatting and generation.}
The retained batched Refusal producer applies the tokenizer's chat
template to a user message with \texttt{tokenize=False} and
\texttt{add\_generation\_prompt=True}, then tokenizes with padding,
truncation, and maximum input length 64. Its ANSWER index is computed
as $\sum_t\mathrm{attention\_mask}_{it}-1$.
The template call does not pass an explicit
\texttt{enable\_thinking} argument. The fixed nominal-dose readout
measurements record right padding and right truncation.
The short-generation probe uses the same template-call arguments
and maximum input length 64 with truncation enabled,
but processes one harmful prompt at a time without padding. At each step it applies the saved nominal-dose patch
at the current final valid input position, with fixed support,
coefficients, and fitting-split mean directions. It generates exactly
sixteen argmax tokens, without EOS-based early termination.
The generation configuration is provided in
\path{configs/refusal_generation.json}, with execution instructions
in \path{docs/experiments.md}.
The batched lexical metrics use the implemented ANSWER index;
the unpadded generation probe directly uses the final valid position.

\paragraph{Operating criteria and statistical units.}
Signed reversal uses a distance ratio of at least two. Dose response
uses signed Spearman correlation $\rho\geq0.95$, with strong negative
ordering $\rho\leq-0.95$ reported separately. Feedback uses the
task-specific endpoint tolerances above. Each feedback observation is
a seed--target batch-mean endpoint; initial zero-dose hits are included
in the total and reported separately.

The independent preservation panels report both a seed mean over
32 prompts and an all-prompt rule, each requiring KL at most 0.001.
The same 32 prompts are reused across eight seeds, giving 256
prompt--seed measurements per selector. For sorted pooled values
$z_{(1)}\leq\cdots\leq z_{(256)}$, the nearest-rank percentile is
\[
P_{95}=z_{(\lceil0.95\times256\rceil)}=z_{(244)}.
\]
The panel measurements use nominal dose and the saved native supports,
coefficients, and direction rules in Appendix~\ref{app:preservation-results}.
These axis-specific operating criteria are separate
from the joint nonlinear certificate in Definition~\ref{def:certificate}.

\begin{table}[!htb]
\caption{\textbf{Formal reference settings and recorded IOI recurrence.}
Setting type distinguishes the formal reference from an executed
analysis. Repeated MCQA/Refusal feedback parameters are specified
only in the executed-settings paragraph above; the clipped controller
of Eq.~\eqref{eq:controller-app} is a reference rule.}
\label{tab:hyper}
\centering
\small
\setlength{\tabcolsep}{5pt}
\begin{tabular}{@{}p{0.14\linewidth}p{0.20\linewidth}p{0.58\linewidth}@{}}
\toprule
Setting type & Component & Setting \\
\midrule
Reference & Refit and certificate & Ridge and preservation weights
$\lambda=\mu=1$; activation-energy bound $E\leq\kappa=2$;
target criterion $\mathrm{Rel}_{\rm tar}(1)\leq0.4$;
preservation bound $L_{\rm off}\leq0.001$. \\
Reference & Sparse selection & \DriverOMP\ residual tolerance
$\varepsilon_{\rm lin}=0.05$ and improvement tolerance
$\eta_{\rm tol}=10^{-3}$; \CoverageDriver\ coverage $\tau=0.95$.
Denominator stabilizers $\delta_0=\eta_0=10^{-3}$; normalized
marginal-gain threshold $g_{\min}=10^{-3}$.
Uniform row weights $w_j=1/n$ for $j=1,\ldots,nr$ give
$W_{\rm row}=\sum_jw_j=r$; Algorithm~\ref{alg:rewrite_coverage_driver}
normalizes coverage by $W_{\rm row}$.
Target normalization guard $\xi=10^{-6}$. \\
Executed & IOI recurrence & Five seeds and five budgets
$k\in\{3,5,8,12,16\}$ give 25 seed--budget configurations;
consensus threshold $\pi^\star=0.6$. \\
\bottomrule
\end{tabular}
\end{table}

The reference linear algebra uses PyTorch \texttt{float32}.
The repeated control runs use MPS or CUDA execution profiles.
Available runtime metadata are included with the corresponding measurement records.  The active-gain step in \DriverOMP\ was checked against \texttt{float64}; the only observed differences were isolated tie breaks between candidates with nearly identical scores.  The formal reference procedure in Algorithm~\ref{alg:rewrite_shared_refit} uses $t_E=10^{-3}$ and $t_\nu=10^{-8}$. It returns the feasible upper-endpoint coefficients when $\kappa-e_{\rm hi}\leq t_E\kappa$ or $\nu_{\rm hi}-\nu_{\rm lo}\leq t_\nu$, solving the cached positive-definite ridge system at each multiplier. These tolerances specify the formal procedure and do not describe the plain-ridge refits used in the recorded scalar-control and module-level reuse runs.

\section{Empirical analysis}
\label{app:empirical}

This appendix organizes the quantitative evidence in the same six-axis
order as Sections~\ref{sec:protocol} and~\ref{sec:exp_overview}.
The comparisons retain each experiment's readout, direction information,
and reporting unit.

\subsection{Experimental configurations and reporting units}
\label{app:emp-setup}

Table~\ref{tab:experiment-configurations} distinguishes the experimental
series. IOI and Sentiment use GPT-2 small; the main MCQA studies use
Qwen3-4B, and Refusal uses Qwen3-1.7B. P denotes evaluation-pair
activation contrasts; F denotes fitting-split mean directions.
IOI/Sentiment reversal, STEM dose, and MCQA/Sentiment
preservation measurements use fixed supports and coefficients
without reselection or refitting. Task definitions and executed
settings are given in Appendices~\ref{app:tasks} and~\ref{app:hyper}.

{\small
\setlength{\tabcolsep}{3pt}
\begin{longtable}{@{}>{\raggedright\arraybackslash}p{0.23\linewidth}>{\raggedright\arraybackslash}p{0.37\linewidth}>{\raggedright\arraybackslash}p{0.36\linewidth}@{}}
\caption{Experimental configurations and reporting units. Panel sizes are per seed; a fitting/evaluation split is written as fit/eval. P and F specify direction information, not different sample sizes.}\label{tab:experiment-configurations}\\
\toprule
Series & Readout and panel & Repetitions and reporting unit \\
\midrule
\endfirsthead
\toprule
Series & Readout and panel & Repetitions and reporting unit \\
\midrule
\endhead
IOI nominal & IO--S margin; 64 pairs, same panel & 10 seeds; stacked $R_{\rm clean}$ and $\mathrm{Rel}_{\rm tar}(1)$ \\
MCQA nominal & Four option logits; 8 questions/category, same panel & 3 seeds/category; stacked $R_{\rm clean}$ and $\mathrm{Rel}_{\rm tar}(1)$ \\
MCQA control & Scalar mean-wrong margin; 24 questions/category, 12/12; head library; P/F & 8 seeds; $8\times6=48$ feedback endpoints/selector/mode; 8 reversal ratios; 8 nominal-dose $\mathrm{Rel}_{\rm tar}(1)$ values/selector/mode \\
Refusal control & Lexical margin; 20 pairs, 10/10; head library; P/F & 8 seeds; $8\times5=40$ feedback endpoints/selector/mode; 8 reversal ratios; 8 nominal-dose $\mathrm{Rel}_{\rm tar}(1)$ values/selector/mode \\
IOI reversal & Saved nominal interventions; same 64 pairs; P & 10 seed-level ratios/selector \\
Sentiment reversal & Saved $k=8$ paired interventions; 16 pairs, 8/8; P/F & 8 seed-level ratios/selector/mode; 8 nominal-dose $\mathrm{Rel}_{\rm tar}(1)$ values/selector/mode \\
Independent dose & Humanities, STEM, Refusal; control evaluation splits; P/F & 8 seeds, 10 doses; 48 curves and 480 points/task \\
Preservation & Fixed 32-prompt MMLU panel; MCQA P, Sentiment/Refusal F; nominal dose & 8 seeds; 256 prompt--seed observations/selector on 32 unique prompts \\
MCQA module reuse & Signed margin recovery; 32 questions/category, 16/16 & 4 seeds/category; evaluation-side coefficient refit or fully frozen \\
Sentiment reuse & Signed margin recovery; 16 pairs, 8/8; P/F; $k=3,5,8$ & 8 seeds; fixed-dose recovery \\
IOI reuse & Pointwise clean-normalized recovery; 64 fitting/64 test pairs; $k\leq8$ & 3 seeds; refit or fixed coefficients, both with evaluation-pair directions \\
IOI dose & 64 pairs, same panel; mean-margin progress and non-answer-position \(\Delta\)KL & Driver-OMP: 2 seeds, $k=4$; Random: 3 seeds, $k=8$ \\
Sentiment unpaired dose & Word-set margin; unpaired prompt pools; seven doses & 3 seeds/method; progress and same-prompt KL \\
IOI feedback & Batch-mean IO--S margin; same 64-prompt panel & 3 seeds, 8 targets; 24 endpoints \\
IOI supporting analyses & Recurrence: $k=3,5,8,12,16$. Backup: separate 64-prompt panel & Recurrence: 5 seeds, 25 configurations. Backup: one to seven sites \\
Model comparison & Four Humanities questions, same panel; four option logits & Qwen3-8B: 5 seeds, residual-only; Mistral-7B: 3 seeds, three module types \\
Refusal continuations & 10 harmful evaluation prompts/seed; 16 greedy tokens; unpatched + 3 selectors & 8 seeds; 320 continuations on 20 reused prompt identities \\
\bottomrule
\end{longtable}
}

Nominal reconstruction, formal reachability, and reuse report means and
sample SD across seeds. Feedback observations are seed--target batch-mean endpoints;
reversal ratios and dose correlations are seed-level quantities.
Preservation retains both prompt and seed identities: 256 measurements
on a reused 32-prompt panel are not 256 different prompts. Continuation
counts likewise retain seed, method, and prompt identities.

\subsection{Results across the six control axes}
\label{app:rq2}

\subsubsection{Reachability: same-panel reconstruction and held-out nominal-dose control}
\label{app:reachability-results}

Table~\ref{tab:nominal-restored} compares five methods on the same
fitting and evaluation panels, using ten IOI seeds and three seeds
per MCQA category. The stacked readouts are IO--S margins for IOI
and four option logits for module-level MCQA. Target recovery is
evaluated using the displacement-normalized residual
\[
\mathrm{Rel}_{\mathrm{tar}}(1)=
\frac{\|B_{\mathrm{target}}-B_{\mathrm{patched}}\|_2}
{\|B_{\mathrm{target}}-B_{\mathrm{source}}\|_2+10^{-6}}.
\]
The complete stacked readout vectors are evaluated within each seed
before computing the across-seed mean, sample standard deviation,
and pass count at $\varepsilon=0.4$. The maximum support budget is
eight. Supports, coefficients, and direction rules are held fixed
at nominal dose $\gamma=1$; the results do not use feedback-selected
doses. Scalar head-level MCQA control uses a separate
correct-minus-mean-wrong margin with the original reference answer
index. Nonlinear greedy obtains the lowest mean residual in all
three same-panel tasks, at additional model-evaluation cost.

Table~\ref{tab:formal_reachability_samepanel} gives the same-panel
residuals and threshold pass counts. For IOI Driver-OMP, the residual
is $0.201\pm0.058$ and passes in 10/10 seeds. Coverage-Driver passes
in 3/3 module-level seeds for both MCQA categories; Humanities
Top-$k$ passes in 2/3 seeds. These results use the fitting panels
and do not by themselves establish held-out generalization.

Table~\ref{tab:nominal-clean} reports the complementary
clean-output-normalized reconstruction error
\[
R_{\rm clean}=
\frac{\|B_{\rm clean}-B_{\rm patched}\|_2}
{\max(\|B_{\rm clean}\|_2,10^{-12})}.
\]
Here $B_{\rm clean}=B_{\rm target}$: both statistics measure the same
reconstruction error, but use different normalization denominators.
The clean-output-normalized scores do not directly determine the
target-residual reachability criterion.

Table~\ref{tab:formal_reachability_heldout} reports the separate held-out fixed-dose experiments, with evaluation-pair directions (P) and training-mean directions (F). On Humanities and STEM, Coverage-Driver and Top-$k$ pass in 8/8 P seeds per category, but in 0/8 F seeds. Driver-OMP passes in 6/8 P seeds in each category and 0/8 F seeds. For Sentiment, Coverage-Driver and Top-$k$ pass in 8/8 P seeds and 0/8 F seeds; Driver-OMP passes in 3/8 P seeds and 0/8 F seeds. Refusal passes in 0/8 seeds for all selectors in either mode. Evaluation-pair directions depend on paired evaluation inputs and are not fully frozen transferable interventions. All counts concern only the target-residual component of the control specification, not the joint nonlinear certificate.

\begin{table}[!htbp]
\centering
\small
\caption{Formal $\mathrm{Rel}_{\mathrm{tar}}(1)$ on the same fitting/evaluation panel at nominal dose $\gamma=1$ and maximum support budget eight. Cells give mean $\pm$ sample SD (number of seeds satisfying $\varepsilon=0.4$/total). IOI uses 64 paired logit-margin readouts (10 seeds); MCQA uses four option logits on eight questions (3 seeds/category). All directions are evaluation-pair (P); lower residuals are better.}
\label{tab:formal_reachability_samepanel}
\setlength{\tabcolsep}{3pt}
\begin{tabular}{@{}lccc@{}}
\toprule
Method & IOI (10 seeds) & Humanities (3 seeds) & STEM (3 seeds) \\
\midrule
Driver-OMP       & $0.201\pm0.058$ (10/10) & $0.279\pm0.032$ (3/3) & $0.175\pm0.035$ (3/3) \\
Coverage-Driver  & $0.294\pm0.029$ (10/10) & $0.152\pm0.154$ (3/3) & $0.061\pm0.033$ (3/3) \\
Top-$k$            & $0.240\pm0.035$ (10/10) & $0.333\pm0.117$ (2/3) & $0.221\pm0.046$ (3/3) \\
Random           & $0.747\pm0.240$ (1/10)  & $1.209\pm0.949$ (0/3) & $1.167\pm1.141$ (0/3) \\
Nonlinear greedy & $0.151\pm0.037$ (10/10) & $0.073\pm0.093$ (3/3) & $0.020\pm0.012$ (3/3) \\
\bottomrule
\end{tabular}
\end{table}

\begin{table}[!htbp]
\centering
\small
\caption{Formal $\mathrm{Rel}_{\mathrm{tar}}(1)$ on held-out evaluation prompts at fixed dose $\gamma=1$ and maximum support budget eight. The unit is one seed-level stacked vector; entries report mean $\pm$ sample SD and number of seeds satisfying $\varepsilon=0.4$ out of eight. Head-level MCQA, paired Sentiment, and Refusal use distinct task-specific readouts. P uses evaluation-pair directions; F uses training-mean directions.}
\label{tab:formal_reachability_heldout}
\setlength{\tabcolsep}{3pt}
\begin{tabular}{@{}llcc@{}}
\toprule
Task & Method & P: mean $\pm$ SD (pass) & F: mean $\pm$ SD (pass) \\
\midrule
Humanities & Driver-OMP      & $0.495\pm0.428$ (6/8) & $1.009\pm0.048$ (0/8) \\
           & Coverage-Driver & $0.188\pm0.072$ (8/8) & $1.012\pm0.058$ (0/8) \\
           & Top-$k$           & $0.192\pm0.071$ (8/8) & $1.010\pm0.055$ (0/8) \\
STEM       & Driver-OMP      & $0.497\pm0.439$ (6/8) & $1.016\pm0.048$ (0/8) \\
           & Coverage-Driver & $0.169\pm0.067$ (8/8) & $1.031\pm0.053$ (0/8) \\
           & Top-$k$           & $0.170\pm0.046$ (8/8) & $1.030\pm0.051$ (0/8) \\
Sentiment  & Driver-OMP      & $0.445\pm0.088$ (3/8) & $0.766\pm0.065$ (0/8) \\
           & Coverage-Driver & $0.171\pm0.084$ (8/8) & $0.746\pm0.056$ (0/8) \\
           & Top-$k$           & $0.167\pm0.084$ (8/8) & $0.742\pm0.052$ (0/8) \\
Refusal    & Driver-OMP      & $0.932\pm0.045$ (0/8) & $0.928\pm0.053$ (0/8) \\
           & Coverage-Driver & $0.799\pm0.037$ (0/8) & $0.801\pm0.039$ (0/8) \\
           & Top-$k$           & $0.774\pm0.043$ (0/8) & $0.779\pm0.044$ (0/8) \\
\bottomrule
\end{tabular}
\end{table}

\begin{table}[!htbp]
\caption{\textbf{Clean-output-normalized reconstruction.} Mean \(\pm\)
sample SD of \(R_{\rm clean}\), lower is better. Same-panel fitting
and evaluation use 64 IOI pairs or eight MCQA questions per category.
Evaluation-pair directions and nominal dose $\gamma=1$ are used.
Maximum budget is eight; early stopping can use fewer candidates.
Nonlinear greedy uses additional model evaluations.}
\label{tab:nominal-clean}
\centering\small
\begin{tabular}{lccc}
\toprule
Method & IOI (10 seeds) & Humanities (3 seeds) & STEM (3 seeds) \\
\midrule
Driver-OMP & $0.386 \pm 0.114$ & $0.045 \pm 0.013$ & $0.030 \pm 0.008$ \\
Coverage-Driver & $0.567 \pm 0.059$ & $0.027 \pm 0.031$ & $0.010 \pm 0.005$ \\
Top-$k$ & $0.464 \pm 0.078$ & $0.054 \pm 0.024$ & $0.038 \pm 0.012$ \\
Random & $1.430 \pm 0.442$ & $0.185 \pm 0.132$ & $0.196 \pm 0.185$ \\
Nonlinear greedy & $0.289 \pm 0.068$ & $0.013 \pm 0.018$ & $0.003 \pm 0.002$ \\
\bottomrule
\end{tabular}

\end{table}

\clearpage
\subsubsection{Signed reversal}
\label{app:formal-reversal-results}

Table~\ref{tab:formal-reversal} combines the extended MCQA/Refusal
experiments with the IOI/Sentiment measurements.
The patch \(-\beta\) is applied to the target member. A ratio of
target-distance to source-distance of at least two defines the
operating pass. Each ratio uses a full evaluation vector; multiple
feedback targets do not multiply the number of reversal replicates.
IOI has ten P seeds and no added F test. Sentiment has eight P and F
seeds, using eight evaluation pairs per seed. The IOI/Sentiment tests keep
the fitted supports and coefficients fixed.

{\small
\setlength{\tabcolsep}{3pt}
\begin{longtable}{lllcl}
\caption{Formal target-side reversal. Ratio at least two is the operating criterion; mean and sample SD are over seeds, not feedback targets. IOI is P-only; P and F are defined in Table~\ref{tab:experiment-configurations}.}\label{tab:formal-reversal}\\
\toprule
Task & Mode & Selector & Pass/seeds & Ratio mean $\pm$ SD \\
\midrule
\endfirsthead
\toprule
Task & Mode & Selector & Pass/seeds & Ratio mean $\pm$ SD \\
\midrule
\endhead
IOI & P & \DriverOMP{} & 10/10 & \(4.7425\pm1.1143\) \\
IOI & P & \CoverageDriver{} & 10/10 & \(2.5702\pm0.3689\) \\
IOI & P & Top-\(k\) & 10/10 & \(3.4315\pm0.8456\) \\
Sentiment & P & \DriverOMP{} & 4/8 & \(2.3722\pm1.0708\) \\
Sentiment & F & \DriverOMP{} & 0/8 & \(0.6547\pm0.2871\) \\
Sentiment & P & \CoverageDriver{} & 8/8 & \(8.5143\pm2.9573\) \\
Sentiment & F & \CoverageDriver{} & 0/8 & \(0.9969\pm0.4172\) \\
Sentiment & P & Top-\(k\) & 8/8 & \(8.7627\pm3.4087\) \\
Sentiment & F & Top-\(k\) & 0/8 & \(1.0000\pm0.4175\) \\
Humanities & P & \DriverOMP{} & 5/8 & \(3.0596\pm1.7541\) \\
Humanities & F & \DriverOMP{} & 0/8 & \(0.1304\pm0.0733\) \\
Humanities & P & \CoverageDriver{} & 8/8 & \(3.7768\pm0.8716\) \\
Humanities & F & \CoverageDriver{} & 0/8 & \(0.1428\pm0.0905\) \\
Humanities & P & Top-\(k\) & 8/8 & \(4.2961\pm2.2079\) \\
Humanities & F & Top-\(k\) & 0/8 & \(0.1424\pm0.0880\) \\
STEM & P & \DriverOMP{} & 7/8 & \(3.6450\pm2.0158\) \\
STEM & F & \DriverOMP{} & 0/8 & \(0.1312\pm0.0658\) \\
STEM & P & \CoverageDriver{} & 8/8 & \(4.8023\pm1.4251\) \\
STEM & F & \CoverageDriver{} & 0/8 & \(0.1520\pm0.0533\) \\
STEM & P & Top-\(k\) & 8/8 & \(5.0921\pm1.5127\) \\
STEM & F & Top-\(k\) & 0/8 & \(0.1441\pm0.0516\) \\
Refusal & P & \DriverOMP{} & 0/8 & \(0.0770\pm0.0513\) \\
Refusal & F & \DriverOMP{} & 0/8 & \(0.0713\pm0.0519\) \\
Refusal & P & \CoverageDriver{} & 0/8 & \(0.1123\pm0.0452\) \\
Refusal & F & \CoverageDriver{} & 0/8 & \(0.1252\pm0.0573\) \\
Refusal & P & Top-\(k\) & 0/8 & \(0.1356\pm0.0527\) \\
Refusal & F & Top-\(k\) & 0/8 & \(0.1452\pm0.0606\) \\
\bottomrule
\end{longtable}
}

Across selectors, Humanities and STEM have 21/24 and 23/24 passing
P method--seed cells, respectively, and no passing F cells. Refusal
has 0/24 in both modes, despite its ordered dose responses below.
The per-seed behavioral vectors and ratios remain in the saved records.
Source-side negative-dose scans start from a different input member
and are not substituted for target-side reversal.

\subsubsection{Dose response}
\label{app:dose-results}

Figure~\ref{fig:independent-dose} uses the independent grid
\(\{-1,-0.5,0,0.25,0.5,0.75,1,1.25,1.5,2\}\).
STEM, Humanities, and Refusal use the same dose grid.
Each task has three selectors, two direction conditions, and eight
seeds: 48 curves and 480 measured points. Ordering uses the signed
Spearman coefficient, with \(\rho\geq0.95\) as the positive-response
criterion. Strong negative ordering \(\rho\leq-0.95\) is reported
separately. Supports, coefficients, and direction construction remain
fixed within each scan; no controller chooses the visited doses.

{\small
\setlength{\tabcolsep}{3pt}
\begin{longtable}{lllccrr}
\caption{Signed Spearman ordering over ten independent doses per seed. Positive and strong negative counts are separate; absolute correlations are not a pass criterion.}\label{tab:independent-dose-stats}\\
\toprule
Task & Mode & Selector & $\rho\geq.95$ & $\rho\leq-.95$ & Mean & SD \\
\midrule
\endfirsthead
\multicolumn{7}{c}{\tablename~\thetable\ (continued)}\\
\toprule
Task & Mode & Selector & $\rho\geq.95$ & $\rho\leq-.95$ & Mean & SD \\
\midrule
\endhead
Humanities & P & \DriverOMP{} & 3/8 & 0/8 & 0.5970 & 0.6005 \\
Humanities & P & \CoverageDriver{} & 5/8 & 0/8 & 0.9394 & 0.0987 \\
Humanities & P & Top-\(k\) & 2/8 & 0/8 & 0.8939 & 0.0727 \\
Humanities & F & \DriverOMP{} & 4/8 & 3/8 & 0.2303 & 1.0202 \\
Humanities & F & \CoverageDriver{} & 5/8 & 3/8 & 0.2500 & 1.0351 \\
Humanities & F & Top-\(k\) & 5/8 & 3/8 & 0.2500 & 1.0351 \\
STEM & P & \DriverOMP{} & 4/8 & 0/8 & 0.5076 & 0.6972 \\
STEM & P & \CoverageDriver{} & 6/8 & 0/8 & 0.9742 & 0.0352 \\
STEM & P & Top-\(k\) & 5/8 & 0/8 & 0.9621 & 0.0495 \\
STEM & F & \DriverOMP{} & 4/8 & 4/8 & -0.0015 & 1.0674 \\
STEM & F & \CoverageDriver{} & 4/8 & 4/8 & 0.0000 & 1.0690 \\
STEM & F & Top-\(k\) & 4/8 & 4/8 & 0.0000 & 1.0690 \\
Refusal & P & \DriverOMP{} & 8/8 & 0/8 & 1.0000 & 0.0000 \\
Refusal & P & \CoverageDriver{} & 8/8 & 0/8 & 1.0000 & 0.0000 \\
Refusal & P & Top-\(k\) & 8/8 & 0/8 & 1.0000 & 0.0000 \\
Refusal & F & \DriverOMP{} & 8/8 & 0/8 & 1.0000 & 0.0000 \\
Refusal & F & \CoverageDriver{} & 8/8 & 0/8 & 1.0000 & 0.0000 \\
Refusal & F & Top-\(k\) & 8/8 & 0/8 & 1.0000 & 0.0000 \\
\bottomrule
\end{longtable}
}

Each selector has three strongly negative Humanities F seeds and four
strongly negative STEM F seeds. All Refusal curves have \(\rho=1\),
although formal reversal fails in both modes. Positive ordering of
forward doses and target-side reversibility are empirically distinct.

\paragraph{IOI and Sentiment dose diagnostics.}
\label{app:legacy-dose}
The IOI panels in Figure~\ref{fig:dose} compare Driver-OMP with four
selected sites and seeds 0/1 against random supports of size eight
and seeds 0/1/2. The full IOI dose series also contains a third
Driver-OMP seed with seven sites. The illustration therefore compares
unequal support sizes and seed counts. Progress is computed from
batch-mean margins. The non-answer-position diagnostic subtracts
KL(corrupt $\|$ clean) from KL(patched $\|$ clean) on the same paired
prompts; it is distinct from the independent 32-prompt preservation
panel in Appendix~\ref{app:preservation-results}.

The Sentiment panels summarize three seeds per method at seven doses,
$\{0,0.25,0.5,1,1.5,2,3\}$, from the saved unpaired-dose records.
They report normalized positive-minus-negative mean-margin progress
and KL on the same negative evaluation prompts, separately from the
paired eight-seed fixed-dose comparison and the independent
32-prompt preservation panel. The record files and their producer
configuration are identified in the accompanying experiment archive.

\subsubsection{Feedback controllability}
\label{app:feedback-results}

Table~\ref{tab:feedback} reports terminal target hits,
while Figure~\ref{fig:feedback-trajectories}
illustrates recorded batch-mean tracking trajectories.
Table~\ref{tab:feedback-details} additionally reports terminal mean absolute
errors, recorded readout-evaluation counts, and initial zero-dose
hits.

Within each seed, the selected support, base
coefficients, and direction rule remain fixed across
requested targets; only the dose is adjusted.
The operating tolerances are $0.3$ for IOI and MCQA
and $0.1$ for Refusal.
The MCQA and Refusal controllers use up to twelve
update iterations followed by a final endpoint
evaluation, without dose clipping or gain halving.

{\small
\setlength{\tabcolsep}{3pt}
\begin{longtable}{lllcrrc}
\caption{\textbf{Repeated scalar feedback.}
Terminal target hits, mean absolute tracking error (MAE),
mean recorded readout-evaluation count, and initial zero-dose hits.
Zero-dose readouts may reuse the cached unpatched baseline.
Errors and evaluation counts are averaged over all seed--target
endpoints, including unsuccessful trials. Initial hits are
included in the total hit count and reported separately.}\label{tab:feedback-details}\\
\toprule
Task & Selector & Mode & Hits & MAE & Evals & Initial \\
\midrule
\endfirsthead
\toprule
Task & Selector & Mode & Hits & MAE & Evals & Initial \\
\midrule
\endhead
IOI & \DriverOMP{} & P & 24/24 & 0.1351 & 3.958 & 2/24 \\
Humanities & \DriverOMP{} & P & 38/48 & 1.0026 & 6.125 & 6/48 \\
Humanities & \DriverOMP{} & F & 8/48 & 2.2641 & 11.479 & 6/48 \\
Humanities & \CoverageDriver{} & P & 46/48 & 0.2378 & 4.417 & 6/48 \\
Humanities & \CoverageDriver{} & F & 8/48 & 1.9239 & 11.479 & 6/48 \\
Humanities & Top-\(k\) & P & 45/48 & 0.2679 & 4.583 & 6/48 \\
Humanities & Top-\(k\) & F & 8/48 & 1.8825 & 11.438 & 6/48 \\
STEM & \DriverOMP{} & P & 39/48 & 1.0743 & 5.354 & 3/48 \\
STEM & \DriverOMP{} & F & 8/48 & 2.1875 & 11.896 & 3/48 \\
STEM & \CoverageDriver{} & P & 47/48 & 0.1321 & 3.854 & 3/48 \\
STEM & \CoverageDriver{} & F & 7/48 & 2.4494 & 11.917 & 3/48 \\
STEM & Top-\(k\) & P & 47/48 & 0.1364 & 3.771 & 3/48 \\
STEM & Top-\(k\) & F & 7/48 & 2.4310 & 11.938 & 3/48 \\
Refusal & \DriverOMP{} & P & 12/40 & 0.1885 & 10.100 & 10/40 \\
Refusal & \DriverOMP{} & F & 12/40 & 0.1870 & 10.100 & 10/40 \\
Refusal & \CoverageDriver{} & P & 21/40 & 0.1167 & 9.450 & 10/40 \\
Refusal & \CoverageDriver{} & F & 21/40 & 0.1174 & 9.375 & 10/40 \\
Refusal & Top-\(k\) & P & 21/40 & 0.1084 & 9.225 & 10/40 \\
Refusal & Top-\(k\) & F & 22/40 & 0.1098 & 9.225 & 10/40 \\
\bottomrule
\end{longtable}
}

Figure~\ref{fig:feedback-trajectories}(a)
shows all three IOI seeds, with eight requested
targets per seed on the same 64-prompt fitting
and evaluation panel. All $24/24$ endpoints are
within tolerance, including two initial zero-dose hits.

Panels (b) and (c) compare Humanities seed 0
and the same six target margins under P and F directions,
with $6/6$ and $1/6$ terminal hits, respectively.
Panel (d) illustrates Refusal under F directions
for seed 0, with $2/5$ hits.
These seed-specific trajectories are included
in the aggregate results of Table~\ref{tab:feedback}
and do not represent additional independent trials.

\subsubsection{Off-target preservation}
\label{app:preservation-results}

The fixed panel has sixteen Humanities and sixteen STEM questions.
Its 32 prompt identities are reused across seeds and methods.
At \(\gamma=1\), the MCQA measurements use each off-target
prompt's own clean-minus-option-exchanged activation contrast. Both
baseline and patched forwards use the original unswapped prompt;
the companion supplies the direction only. Inputs are processed
without padding, and the patch and full-vocabulary KL readout use
the final valid input position. Saved task-native supports and
coefficients remain fixed. Sentiment and Refusal broadcast their
fitting-split mean directions; the added Sentiment measurements use
the saved $k=8$ interventions. Consequently,
the MCQA measurement is not a fully frozen-direction intervention
on unseen inputs. These measurements are distinct from
the paired-position diagnostics in Figure~\ref{fig:dose}.

For Refusal, 196/256, 82/256, and 93/256 prompt--seed measurements
satisfy the KL criterion of at most 0.001 for \DriverOMP{},
\CoverageDriver{}, and Top-\(k\), respectively.

{\small
\setlength{\tabcolsep}{3pt}
\begin{longtable}{llrrrrcc}
\caption{Full-vocabulary KL(unpatched $\|$ patched) at nominal dose, using natural logarithms. Each row pools 32 prompts $\times$ eight seeds (256 measurements on 32 unique prompts). P95 is nearest rank. Last columns count seeds meeting mean-KL or all-32-prompt KL $\leq0.001$. MCQA uses each off-prompt pair; Sentiment/Refusal use training means.}\label{tab:fixed-panel-preservation}\\
\toprule
Task & Selector & Mean KL & Median & P95 & Max & Mean-pass & All-pass \\
\midrule
\endfirsthead
\toprule
Task & Selector & Mean KL & Median & P95 & Max & Mean-pass & All-pass \\
\midrule
\endhead
Sentiment & \DriverOMP{} & 0.005527 & 0.004959 & 0.011119 & 0.0286 & 0/8 & 0/8 \\
Sentiment & \CoverageDriver{} & 0.011015 & 0.009627 & 0.024779 & 0.0381 & 0/8 & 0/8 \\
Sentiment & Top-\(k\) & 0.011567 & 0.010322 & 0.024746 & 0.0387 & 0/8 & 0/8 \\
Humanities & \DriverOMP{} & 0.496530 & 0.050394 & 2.349920 & 15.5523 & 0/8 & 0/8 \\
Humanities & \CoverageDriver{} & 0.109692 & 0.050405 & 0.392607 & 1.0154 & 0/8 & 0/8 \\
Humanities & Top-\(k\) & 0.115396 & 0.056724 & 0.478003 & 0.6963 & 0/8 & 0/8 \\
STEM & \DriverOMP{} & 0.581701 & 0.051584 & 3.088410 & 17.7225 & 0/8 & 0/8 \\
STEM & \CoverageDriver{} & 0.128467 & 0.060374 & 0.465265 & 1.0098 & 0/8 & 0/8 \\
STEM & Top-\(k\) & 0.118910 & 0.055203 & 0.511026 & 0.7559 & 0/8 & 0/8 \\
Refusal & \DriverOMP{} & 0.001080 & 0.000174 & 0.005086 & 0.0410 & 4/8 & 2/8 \\
Refusal & \CoverageDriver{} & 0.008658 & 0.002119 & 0.035977 & 0.1698 & 0/8 & 0/8 \\
Refusal & Top-\(k\) & 0.007986 & 0.001948 & 0.036849 & 0.1572 & 0/8 & 0/8 \\
\bottomrule
\end{longtable}
}

For Humanities, STEM, and Sentiment, none of the
eight seeds satisfies either preservation criterion
under any selector. On the Refusal
panel, Driver-OMP passes in four seeds by the mean rule and two by
the all-prompt rule; Coverage-Driver and Top-\(k\) pass in none.
Large predictive-distribution changes are not confined to one extreme
observation: the MCQA medians are approximately
0.05--0.06. The normalized question-string check found no exact
duplicates between the fixed panel and the native MCQA task pools,
but that does not establish semantic independence. These KL results
measure predictive-distribution change, not completed-answer safety
or semantic correctness.

\subsubsection{Held-out reuse}
\label{app:rq4}

The module-level MCQA study uses four seeds per category, 32 questions
split 16/16, and signed mean-margin recovery. Table~\ref{tab:baselines-transfer}
reports this four-seed series separately from the head-level controls. Refit uses evaluation targets and evaluation-pair directions;
fully frozen open-loop reuses the fitting-side support, coefficients,
mean directions, and nominal dose. Recovery can exceed one because
it is a signed margin ratio, not an accuracy or clipped probability.

\begin{table}[!htbp]
\caption{\textbf{Module-level MCQA reuse.} Seed means and
sample SD. Refit retains evaluation-target access; fully frozen means the full
open-loop intervention is fixed. Each category has four seeds.}
\label{tab:baselines-transfer}
\centering\small
\begin{tabular}{llccc}
\toprule
Task & Method & Seeds & Refit-$\beta$ & Fully frozen \\
\midrule
Humanities & Driver-OMP & 4 & $0.992 \pm 0.009$ & $0.021 \pm 0.026$ \\
Humanities & Coverage-Driver & 4 & $0.995 \pm 0.053$ & $0.033 \pm 0.029$ \\
Humanities & Top-$k$ & 4 & $1.078 \pm 0.017$ & $0.028 \pm 0.024$ \\
Humanities & Random & 4 & $0.758 \pm 0.443$ & $0.022 \pm 0.023$ \\
STEM & Driver-OMP & 4 & $0.987 \pm 0.034$ & $0.024 \pm 0.066$ \\
STEM & Coverage-Driver & 4 & $0.944 \pm 0.056$ & $0.038 \pm 0.108$ \\
STEM & Top-$k$ & 4 & $1.075 \pm 0.029$ & $0.031 \pm 0.093$ \\
STEM & Random & 4 & $0.744 \pm 0.424$ & $0.041 \pm 0.108$ \\
\bottomrule
\end{tabular}

\end{table}

The IOI study in Table~\ref{tab:ioi-reuse} uses seeds
$s\in\{0,1,2\}$, a maximum support budget of eight,
and 64 fitting pairs plus 64 test pairs per seed.
The two panels are sampled separately from the same prompt
distribution, using seeds $s$ and $s+10{,}000$, respectively.
For test pair $i$, let $b_i^{\rm clean}$ and $b_i^{\rm patched}$
be the scalar IO--S margins. The pointwise error and seed-level
recovery are
\[
e_i=\frac{|b_i^{\rm patched}-b_i^{\rm clean}|}
{\max(|b_i^{\rm clean}|,10^{-6})},\qquad
\mathrm{Rec}=1-\frac{1}{64}\sum_{i=1}^{64}e_i.
\]
A test pair fails when $e_i>0.3$. Each failure fraction is
$64^{-1}\sum_{i=1}^{64}\mathbf{1}\{e_i>0.3\}$;
the table reports its arithmetic mean across the three seeds.
Recovery is summarized by the mean and sample SD across seeds.
Both coefficient-refit and fixed-coefficient branches use
evaluation-pair clean--corrupt activation differences.
The fixed-coefficient branch is therefore not a fully frozen
intervention, and this pointwise statistic is distinct from
MCQA mean-margin recovery.

\begin{table}[!htbp]
\caption{\textbf{IOI reuse on the test split.}
Recovery is reported as the mean $\pm$ sample SD
across three seeds. Fixed coefficients retain access
to evaluation-pair directions. Failure fractions
use the pointwise error and threshold specified above.}
\label{tab:ioi-reuse}
\centering\small
\begin{tabular}{lcccc}
\toprule
Method & Refit rec. & Fixed-$\beta$ rec. & Refit fail frac. & Fixed-$\beta$ fail frac. \\
\midrule
Driver-OMP & $0.461 \pm 0.144$ & $0.391 \pm 0.111$ & 0.583 & 0.661 \\
Coverage-Driver & $0.449 \pm 0.083$ & $0.298 \pm 0.143$ & 0.667 & 0.833 \\
Top-$k$ & $0.632 \pm 0.012$ & $0.434 \pm 0.160$ & 0.464 & 0.734 \\
Random & $-0.670 \pm 0.612$ & $-0.649 \pm 0.637$ & 0.984 & 0.984 \\
\bottomrule
\end{tabular}

\end{table}

{\small
\setlength{\tabcolsep}{3pt}
\begin{longtable}{lclrrc}
\caption{Sentiment paired fixed-dose recovery: seed mean and sample SD across eight seeds, by support budget and direction condition. Forward mean-margin recovery is distinct from formal vector reversal.}\label{tab:sentiment-reuse}\\
\toprule
Selector & $k$ & Mode & Mean & SD & Seeds \\
\midrule
\endfirsthead
\multicolumn{6}{c}{\tablename~\thetable\ (continued)}\\
\toprule
Selector & $k$ & Mode & Mean & SD & Seeds \\
\midrule
\endhead
\DriverOMP{} & 3 & P & 0.3602 & 0.1887 & 8 \\
\DriverOMP{} & 3 & F & 0.3536 & 0.1723 & 8 \\
\CoverageDriver{} & 3 & P & 0.7687 & 0.1038 & 8 \\
\CoverageDriver{} & 3 & F & 0.8137 & 0.4851 & 8 \\
Top-\(k\) & 3 & P & 0.7804 & 0.1173 & 8 \\
Top-\(k\) & 3 & F & 0.8127 & 0.4852 & 8 \\
Random & 3 & P & 0.0745 & 0.1221 & 8 \\
Random & 3 & F & 0.0743 & 0.1072 & 8 \\
\DriverOMP{} & 5 & P & 0.4949 & 0.1142 & 8 \\
\DriverOMP{} & 5 & F & 0.4538 & 0.1586 & 8 \\
\CoverageDriver{} & 5 & P & 0.9542 & 0.0969 & 8 \\
\CoverageDriver{} & 5 & F & 0.8891 & 0.5155 & 8 \\
Top-\(k\) & 5 & P & 0.9328 & 0.1041 & 8 \\
Top-\(k\) & 5 & F & 0.8880 & 0.5157 & 8 \\
Random & 5 & P & 0.1065 & 0.1376 & 8 \\
Random & 5 & F & 0.0845 & 0.1030 & 8 \\
\DriverOMP{} & 8 & P & 0.5928 & 0.1053 & 8 \\
\DriverOMP{} & 8 & F & 0.5291 & 0.2137 & 8 \\
\CoverageDriver{} & 8 & P & 1.0311 & 0.1120 & 8 \\
\CoverageDriver{} & 8 & F & 0.9420 & 0.5354 & 8 \\
Top-\(k\) & 8 & P & 1.0325 & 0.1085 & 8 \\
Top-\(k\) & 8 & F & 0.9370 & 0.5262 & 8 \\
Random & 8 & P & 0.1532 & 0.1601 & 8 \\
Random & 8 & F & 0.1122 & 0.1246 & 8 \\
\bottomrule
\end{longtable}
}

\paragraph{Interpretation of reuse.}\label{app:heldout-pass}
High refit recovery demonstrates reuse of selected sites with
evaluation-side coefficient refitting. High frozen forward recovery does not establish backward
reversal, preservation, or feedback coverage. The results therefore
retain continuous recovery and their information conditions instead
of converting heterogeneous series into one cross-task pass flag.

The paired Sentiment series can show high mean forward recovery with
training-mean directions, particularly for Coverage-Driver and
Top-\(k\), while formal F reversal fails for every
method and seed. In the MCQA and Refusal feedback experiments, F directions remain
fixed while dose adapts to observed target errors; fully frozen
open-loop reuse also fixes the dose. No common cross-task recovery threshold or correction family
is imposed on these different statistics.

The strong Sentiment F mean-margin recovery reported in the fixed-dose reuse table does not imply low formal target residuals: mean-margin recovery aggregates across prompts, whereas $\mathrm{Rel}_{\mathrm{tar}}$ measures the remaining error in the entire vector of held-out prompt-level readouts. Under the maximum support budget of eight, none of the three evaluated selectors satisfies $\varepsilon=0.4$ in any of the eight Sentiment F seeds.

\subsection{Comparison methods and sensitivity}
\label{app:comparisons-sensitivity}

\subsubsection{Comparison methods}
\label{app:rq3}

The nominal table compares both sparse selectors, Top-\(k\), random
supports, and higher-cost nonlinear greedy using the task-specific
dictionaries and refits. Equal maximum support budgets
do not imply equal realized support sizes after early stopping or
equal model-evaluation cost. In the repeated scalar controls,
Top-\(k\) ranks \(|G_{\rm train}^{\top}q_{\rm train}|\), and
all three selectors fit on the same fitting split.

\paragraph{Random and nonlinear-greedy references.}
Random supports are sampled from the task-specific candidate library
and evaluated using the task-specific direction and refitting rules.
Budget matching holds the maximum selection-cost or cardinality budget
fixed; support-size matching holds the realized number of selected
candidates fixed. The matching condition is specified for each
experimental series. In nominal reconstruction, random supports contain
$\min(k,|\V|)$ candidates at the prescribed cardinality budget $k$;
the other selectors may stop earlier.

Nonlinear greedy provides a higher-cost reconstruction reference.
At each iteration, it shortlists up to 32 remaining candidates by
absolute correlation with the current local residual. Each trial
support is refitted by ridge regression and evaluated in the nonlinear
model; the candidate with the lowest $R_{\rm clean}$ is added.
The procedure stops at its support budget or the implemented
mean-absolute-error criterion. It is a greedy heuristic rather than
a global optimization bound.

The Refusal single-site reference fixes answer head L15.H0.
Its P and F feedback both give 10/40 hits, equal to the ten initial
zero-dose hits; it adds no newly hit endpoints. Its formal reversal
is 0/8 in each mode. At nominal dose, its mean preservation KL is
$5.29\times10^{-7}$ over 256 prompt--seed measurements: very
small distributional change accompanies little target effect. This head-level proxy is not canonical
residual-stream ActAdd \citep{turner2023activation}, CAA
\citep{rimsky2024caa}, or RepE \citep{zou2023representation}.
Systematic comparisons
with those steering families require matched readouts, information
access, budgets, and intervention definitions.

\subsubsection{Sensitivity}
\label{app:rq5}

Formal reachability was separately reclassified at $\varepsilon\in\{0.30,0.35,0.40,0.45,0.50\}$, distinct from the absolute feedback-endpoint tolerances in the existing feedback sensitivity table. IOI Driver-OMP passes in 9/10 seeds at $\varepsilon=0.30$ and 10/10 at $\varepsilon\ge0.35$. On the held-out MCQA and paired Sentiment panels, all training-mean-direction (F) configurations remain at 0/8 seeds throughout this range. Refusal similarly fails in all P and F conditions even at $\varepsilon=0.50$. For Coverage-Driver in P mode, Humanities passes in 8/8 seeds throughout the range, STEM reaches 8/8 at $\varepsilon\ge0.35$, and Sentiment reaches 8/8 at $\varepsilon\ge0.40$. Some individual classifications are sensitive: nominal module-level Humanities Top-$k$ passes in 2/3 seeds at $\varepsilon=0.40$ and 3/3 at $\varepsilon=0.45$. Complete per-condition sensitivity counts are included in the accompanying reachability records.

The sensitivity analyses vary support budget, target-residual threshold,
endpoint tolerance, signed ordering threshold, and preservation
aggregation within each experimental series.

Sentiment's support-budget dependence is retained in
Table~\ref{tab:sentiment-reuse}; signed dose tails are in
Table~\ref{tab:independent-dose-stats}; mean versus all-prompt
preservation rules are in Table~\ref{tab:fixed-panel-preservation}.
Table~\ref{tab:feedback-tolerances} reclassifies the same measured
feedback endpoints at five absolute-error tolerances.

{\small
\setlength{\tabcolsep}{3pt}
\begin{longtable}{lllccccc}
\caption{Feedback endpoint hits reclassified at five absolute-error tolerances. These are the same endpoints, not new runs or independent samples.}\label{tab:feedback-tolerances}\\
\toprule
Task & Selector & Mode & 0.10 & 0.20 & 0.30 & 0.40 & 0.50 \\
\midrule
\endfirsthead
\toprule
Task & Selector & Mode & 0.10 & 0.20 & 0.30 & 0.40 & 0.50 \\
\midrule
\endhead
Humanities & \CoverageDriver{} & F & 1/48 & 2/48 & 8/48 & 9/48 & 10/48 \\
Humanities & \CoverageDriver{} & P & 11/48 & 24/48 & 46/48 & 46/48 & 46/48 \\
Humanities & \DriverOMP{} & F & 1/48 & 2/48 & 8/48 & 8/48 & 10/48 \\
Humanities & \DriverOMP{} & P & 6/48 & 17/48 & 38/48 & 38/48 & 38/48 \\
Humanities & Top-\(k\) & F & 1/48 & 2/48 & 8/48 & 10/48 & 10/48 \\
Humanities & Top-\(k\) & P & 10/48 & 29/48 & 45/48 & 45/48 & 45/48 \\
STEM & \CoverageDriver{} & F & 1/48 & 2/48 & 7/48 & 10/48 & 13/48 \\
STEM & \CoverageDriver{} & P & 19/48 & 39/48 & 47/48 & 48/48 & 48/48 \\
STEM & \DriverOMP{} & F & 1/48 & 2/48 & 8/48 & 9/48 & 11/48 \\
STEM & \DriverOMP{} & P & 15/48 & 29/48 & 39/48 & 39/48 & 39/48 \\
STEM & Top-\(k\) & F & 1/48 & 2/48 & 7/48 & 10/48 & 12/48 \\
STEM & Top-\(k\) & P & 21/48 & 35/48 & 47/48 & 47/48 & 47/48 \\
Refusal & \CoverageDriver{} & F & 21/40 & 37/40 & 40/40 & 40/40 & 40/40 \\
Refusal & \CoverageDriver{} & P & 21/40 & 37/40 & 40/40 & 40/40 & 40/40 \\
Refusal & \DriverOMP{} & F & 12/40 & 22/40 & 33/40 & 38/40 & 39/40 \\
Refusal & \DriverOMP{} & P & 12/40 & 22/40 & 32/40 & 38/40 & 39/40 \\
Refusal & Top-\(k\) & F & 22/40 & 39/40 & 40/40 & 40/40 & 40/40 \\
Refusal & Top-\(k\) & P & 21/40 & 39/40 & 40/40 & 40/40 & 40/40 \\
\bottomrule
\end{longtable}
}

At tolerance 0.50, the three Humanities F methods reach only 10/48
endpoints each, and the STEM F methods reach 11/48, 13/48, and 12/48
for \DriverOMP{}, \CoverageDriver{}, and Top-\(k\), respectively.
The MCQA P/F difference therefore persists across the reported
0.10--0.50 range. Refusal's 48 independent dose curves all have
$\rho=1$, so its positive-ordering counts remain unchanged for signed
thresholds from 0.80 to 0.99. Strongly negative MCQA curves are not
counted as positive responses at any of these thresholds.

\subsection{Supporting analyses}
\label{app:supporting-results}

\subsubsection{IOI selection recurrence and backup augmentation}
\label{app:rq1}

The recurrence analysis crosses five random seeds with five budgets
\(k\in\{3,5,8,12,16\}\), giving 25 configurations. Two late
head--position candidates, L9.END.H9 and L10.END.H7, each appear in
15/25 configurations. The configured recurrence threshold is 0.6.
These frequencies describe this seed--budget grid; they are not
bootstrap probabilities or evidence that either site is necessary.
Their locations can be compared with the manual IOI circuit of
\citet{wang2022interpretability}.

\begin{figure}[!htbp]
\centering
\includegraphics[width=0.95\linewidth]{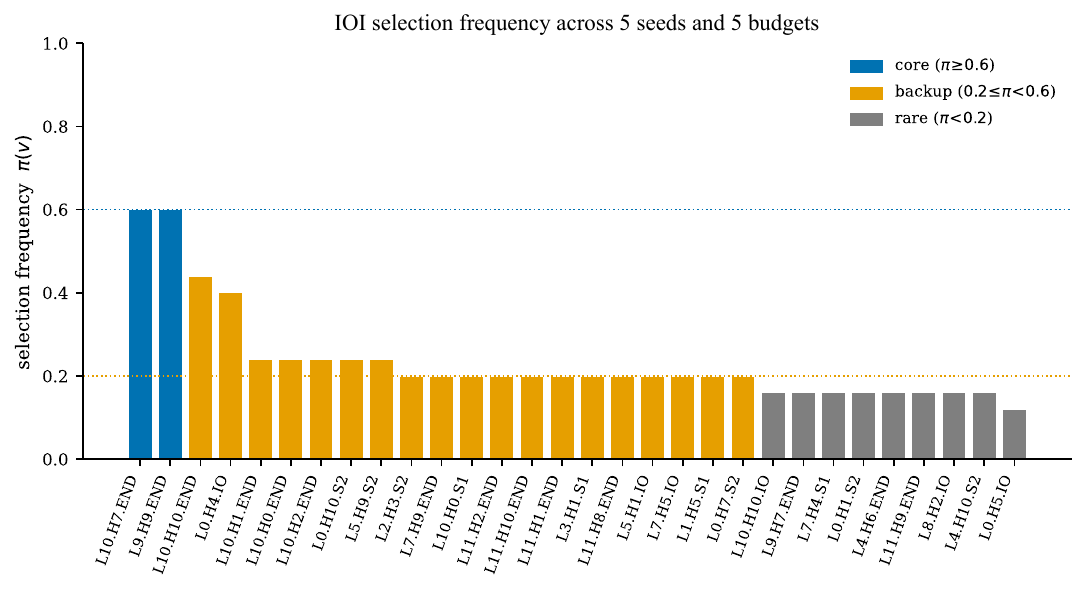}
\caption{\textbf{IOI selection recurrence.} Frequencies across five
seeds and five budgets. Two candidates occur in 15/25 configurations;
the labels ``core'', ``backup'', and ``rare'' denote recurrence bands
in this analysis, not proofs of necessity or universal substitutability.}
\label{fig:driver-class-freq-app}
\end{figure}

The separate backup experiment starts from L9.END.H9 and adds up to
six candidates. Across 64 prompts, mean pointwise absolute error
normalized by $\max(|b_{\rm clean}|,10^{-6})$ changes from 0.5894
with one site to 0.3503 with seven. The first addition worsens the mean
to 0.698. At threshold 0.3, 26 prompts cross toward smaller error,
six cross in the opposite direction, and 29/64 still fail at the final
support. This statistic differs from stacked-vector $R_{\rm clean}$,
and the six-site random reference is not size-matched to the final
seven-site support. The complete support-by-prompt record is retained
with the experimental data.

\subsubsection{Model scope}
\label{app:model-scope}

Table~\ref{tab:source-bound-scale} compares nominal reconstruction across three model
configurations. All use four option logits
and clean-output vector normalization, but Qwen3-4B uses eight
questions per seed and three module types; Qwen3-8B uses four
questions and residual-only candidates; Mistral uses four questions
and three module types. They have three, five, and three seeds,
respectively. Their differences cannot isolate the causal effect of
model size or architecture.

\begin{table}[!htbp]
\caption{\textbf{Configuration-specific model comparisons.} Mean \(\pm\)
sample SD of nominal \(R_{\rm clean}\). Panels and libraries differ
as specified in the text; these are configuration-specific results.}
\label{tab:source-bound-scale}
\centering\small
\begin{tabular}{lccccc}
\toprule
Model & Seeds & Driver-OMP & \shortstack{Coverage-\\Driver} & Top-$k$ & Random \\
\midrule
Qwen3-4B & 3 & $0.045 \pm 0.013$ & $0.027 \pm 0.031$ & $0.054 \pm 0.024$ & $0.185 \pm 0.132$ \\
Qwen3-8B & 5 & $0.016 \pm 0.008$ & $0.018 \pm 0.009$ & $0.027 \pm 0.012$ & $0.077 \pm 0.032$ \\
Mistral-7B & 3 & $0.044 \pm 0.005$ & $0.015 \pm 0.003$ & $0.008 \pm 0.006$ & $0.050 \pm 0.047$ \\
\bottomrule
\end{tabular}

\end{table}

\subsubsection{Refusal short continuations}
\label{app:rq6}
\label{app:short-continuations}

The Refusal probe has eight seeds, ten harmful evaluation
prompts per seed, and four conditions (unpatched plus three selectors),
giving 320 continuations on twenty distinct prompt identities reused
across seeds. All continuations have exactly sixteen greedy tokens,
begin with \texttt{<think>}, and contain no \texttt{</think>} marker.
The observed window therefore does not expose completed final answers.
It cannot determine final semantic refusal or compliance rates.
Patches use fixed training-mean directions at each current last valid
position. This unpadded generation procedure is separate from the
batched lexical-readout configuration in Appendix~\ref{app:hyper}.

\subsection{Limitations}
\label{app:limitations}

The experiments cover a finite set of models, task panels, readouts,
and candidate libraries. Evaluation-pair directions require activation
contrasts from paired evaluation inputs, so their strong MCQA results
do not establish control without this information. Training-mean
feedback still adapts dose at evaluation time, whereas fully frozen
open-loop reuse fixes the entire intervention. The executed MCQA and
Refusal plain-ridge fits impose neither an independent physical-energy
bound nor an off-target penalty, and separate axis results do not
establish the joint nonlinear certificate. Preservation is measured on
32 reused MMLU prompts; it quantifies predictive-distribution change
rather than semantic correctness or safety. Refusal's lexical readout
and sixteen-token continuations leave completed-answer behavior
unresolved. Sentiment feedback controllability remains unassessed
because the reported Sentiment series contain no multi-target feedback
trajectories. Different model panels and candidate
libraries preclude an isolated scaling comparison, while the single-site
proxy does not replace matched implementations of established steering
methods. Broader control and transfer claims require corresponding
evaluations beyond these settings.

\section{Reproducibility and Statistical Reporting}
\label{app:reproducibility}

\paragraph{Code and experimental records.}
Code and experimental records are publicly available at
\url{https://github.com/zxzok/sparse-internal-control}.
The repository contains the implementation,
experiment configurations, evaluation inputs, saved interventions,
task-specific readout vectors, prompt-level KL measurements,
and analysis scripts for reproducing the reported tables and statistics.
Experiment identities and sample configurations
are indexed in Table~\ref{tab:experiment-configurations}.

\paragraph{Models and execution environments.}
The experiments use GPT-2 small, Qwen3-1.7B, Qwen3-4B,
and the additional models specified in Appendix~\ref{app:model-scope}.
MCQA nominal reconstruction and the documented model-scale
runs used an Apple M2 Ultra system with the MPS backend.
The IOI/Sentiment reversal experiments, Sentiment and MCQA
preservation measurements, and independent STEM dose scan
used NVIDIA RTX A6000 GPUs with CUDA.
Series-specific execution profiles and available runtime metadata
are documented in \path{experiments/execution_profiles.json},
\path{docs/compute_resources.md}, and the corresponding measurement records.
Batched Refusal lexical scores use the ANSWER index
$\sum_t\mathrm{attention\_mask}_{it}-1$; this index coincides with the
last valid input token under right padding. The short-generation
probe uses unpadded inputs and the actual final valid token.
Available model repository identifiers and checkpoint revisions are
listed in \path{experiments/model_assets.json}.
Software dependencies and execution profiles are documented in
\path{requirements/model-*.lock.txt} and
\path{experiments/execution_profiles.json}, respectively.

\paragraph{Statistical analysis.}
Nominal reconstruction, formal reachability, and held-out reuse report
means and sample standard deviations across seeds.
Feedback results use seed--target endpoints, whereas
signed reversal and dose response use seed-level
ratios and curves. Preservation statistics retain
both prompt and seed identities. 
No common multiplicity correction is applied across the six control axes.

\end{document}